\documentclass{article}

\usepackage{iclr2027_conference,times}

\usepackage[utf8]{inputenc} % allow utf-8 input
\usepackage[T1]{fontenc}    % use 8-bit T1 fonts
\usepackage{hyperref}       % hyperlinks
\usepackage{xurl}           % allow long bibliography URLs to wrap
\usepackage{booktabs}       % professional-quality tables
\usepackage{amsfonts}       % blackboard math symbols
\usepackage{nicefrac}       % compact symbols for 1/2, etc.
\usepackage{microtype}      % microtypography
\usepackage{xcolor}         % colors
\usepackage{amsmath}
\usepackage{float}
\usepackage{needspace}
\usepackage{subcaption}
\usepackage{graphicx}
\usepackage{array}

\title{Existence-Field Diffusion Model for Spatial Point Processes with Variable Cardinality}

\author{%
  \begin{tabular}{@{}p{0.48\textwidth}@{\hspace{0.04\textwidth}}p{0.48\textwidth}@{}}
    \textbf{Xiaoyin Pan} & \textbf{Christian R. Shelton} \\
    University of California, Riverside & University of California, Riverside \\
    \texttt{xpan041@ucr.edu} & \texttt{cshelton@cs.ucr.edu} \\[1.2em]
    \textbf{Rakshith Mahishi} & \textbf{Chengkuan Hong} \\
    New York University & Zhongguancun Academy \\
    \texttt{rm6801@nyu.edu} & \texttt{hongchengkuan@bza.edu.cn}
  \end{tabular}%
}

\iclrfinalcopy
\makeatletter
\renewcommand{\@maketitle}{\vbox{\hsize\textwidth
  \lhead{}
  {\LARGE\scshape\@title\par}
  \vskip 1.2em
  {\normalfont\noindent\@author\par}
  \vskip 0.3in minus 0.1in
}}
\makeatother
\renewcommand{\headrulewidth}{0pt}
\hypersetup{
  hidelinks,
  pdftitle={Existence-Field Diffusion Model for Spatial Point Processes with Variable Cardinality},
  pdfauthor={Xiaoyin Pan; Christian R. Shelton; Rakshith Mahishi; Chengkuan Hong},
  pdfsubject={Spatial point processes with variable cardinality},
  pdfkeywords={diffusion models, spatial point processes, variable cardinality}
}

\begin{document}
	\maketitle
	\begin{abstract}
We study generative modeling of spatial point processes (SPP), where both the number of points and their spatial configuration are governed by a joint distribution.
While diffusion models have achieved strong performance in modeling complex distributions, extending them to variable-cardinality SPP remains challenging. 
Existing approaches either sample cardinality before generating locations conditionally, or introduce specialized discrete operations to change the number of points during generation.
We propose the existence-field diffusion model (EFDM), which associates each potential point with a continuous variable representing its degree of presence.
EFDM uses Gaussian diffusion to jointly model point locations and cardinality through a simple continuous representation.
Existence probabilities can increase or decrease in both forward and reverse diffusion, without explicit point-addition or point-deletion steps.
The same construction naturally accommodates categorical attributes, allowing locations, existence, and attributes to be modeled together.
Experiments demonstrate improved molecular stability and validity over the compared baselines on the molecular dataset, alongside competitive performance on real-world spatial and synthetic datasets.
\end{abstract}

	\section{Introduction}
Diffusion models \citep{ho2020denoising,song2020score,sohl2015deep} have achieved success in generative modeling, enabling high-quality generation in many domains.
In applications such as image and audio generation, these models commonly
operate on ordered arrays of a prescribed size.
However, many real-world data are naturally represented as point sets with variable cardinality, such as spatial events, molecular structures, and particle systems. 
Modeling such data requires handling unordered point sets and learning distributions over both the number of points and their spatial configuration. Distributions over such objects are called spatial point processes (SPP).
Diffusion models for SPPs pose a fundamental challenge: how to model both cardinality and spatial structure in a unified and permutation-invariant manner.

Real-world point sets exhibit dependencies among points as well as
between cardinality and spatial configuration. In a molecule, bond
lengths, bond angles, and steric constraints couple atomic positions:
placing one atom restricts where others can lie, and all atoms must
form a chemically consistent structure. Changing the number of atoms
also changes the feasible molecular structures. In transportation data,
point sets might represent e-scooter pick-up locations. Shared fluctuations in
travel demand can produce spatial clusters, making
observed locations informative about where other events are likely to
occur and how many are expected. These examples motivate modeling the
entire point set jointly, capturing both dependencies among points and
their relationship with cardinality.

Point-process models capture event dependencies through conditional
intensities~\citep{hawkes1971spectra,mei2017neural,chen2020neural}
or latent processes~\citep{hong2022deep}.
Diffusion models have also been applied to 3D shape generation with
point clouds~\citep{luo2021diffusion,ji2024latent}.
In these tasks, point locations describe the shape's geometry, while
the point count specifies its sampling resolution. For spatial events,
molecules, and particle systems, cardinality is itself part of the
random configuration to be modeled.
To handle variable cardinality, a fixed-dimensional diffusion model
(FDDM)~\citep{hoogeboom2022equivariant} first samples a point count and
then generates locations conditioned on that count.
In particle physics, \citet{mikuni2023fast} likewise first generate a
particle count and properties of the collection as a whole, then
conditionally generate the particles' features.
These hierarchical approaches can capture dependencies between cardinality
and spatial configuration, but keep the sampled count fixed during
point-level diffusion.
In contrast, the trans-dimensional diffusion model
(TDDM)~\citep{campbell2023trans} and Point Set Diffusion
(PSD)~\citep{ludke2024unlocking} jointly generate cardinality and locations,
allowing both to evolve within a single generative process.
However, they change point presence through discrete additions and
deletions, rather than refining it gradually together with point
locations.

\begin{figure}[t]
	\centering
	\makebox[\textwidth][c]{
		\includegraphics[width=\textwidth]{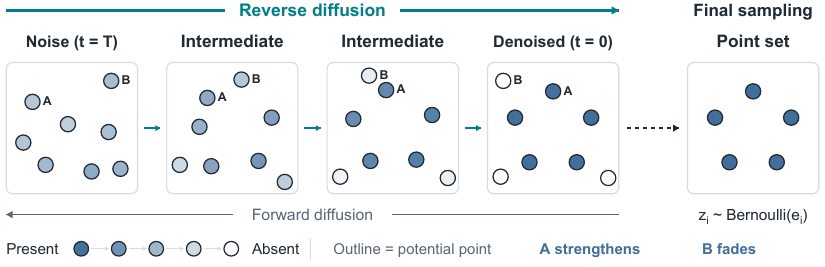}
	}
	\caption{
		Binary EFDM jointly diffuses positions and existence.
		Outlines mark potential points; fill shading qualitatively indicates the degree of existence.
		In reverse diffusion, A strengthens and B fades.
		Final Bernoulli sampling yields the point set. Paths are schematic.
	}
	\label{fig:forward_reverse_sketch}
\end{figure}

We propose the existence-field diffusion model (EFDM), which brings
variable-cardinality point sets into a Gaussian diffusion framework.
Each potential point has a continuous existence variable that is
diffused together with its coordinates (Fig.~\ref{fig:forward_reverse_sketch}).
A final readout converts the recovered fields into a
discrete point set. Extending the existence field with categorical
states also allows discrete attributes to be modeled in the same
diffusion process. EFDM jointly refines point presence, geometry, and categorical attributes through Gaussian diffusion, without fixing cardinality before denoising as in FDDM or introducing explicit point additions and deletions as in TDDM and PSD.

On the molecular dataset QM9, EFDM jointly generates molecular geometry
and categorical atom types with superior stability and validity
relative to the reported FDDM and TDDM baselines. On real-world and synthetic datasets
without point attributes, EFDM achieves competitive scores on joint
count--configuration metrics. These comparisons provide empirical
evidence of its ability to model such dependencies.

\section{Related Work}
\label{sec:related_work}

\paragraph{Temporal and Spatial Point Processes.}
Existing point-process methods describe event distributions through
intensity functions or latent processes. First-order intensity functions
describe the expected local density of events but do not, by themselves,
characterize interactions or determine the likelihood of an entire point
configuration~\citep{daley2003introduction}.
In temporal point processes, history-dependent conditional intensities
capture event interactions and determine event-sequence likelihoods;
neural models parameterize them with recurrent networks~\citep{mei2017neural}
or model event sequences with continuous-time normalizing
flows~\citep{chen2020neural}.
Other approaches introduce latent structure: \citet{zhou2022neural} use
latent dynamics to parameterize an intensity with closed-form integration,
while \citet{hong2022deep} build a hierarchy of latent Poisson processes.

\paragraph{Point Set Generation.}
For geometric data, existing point-set generators use a variety of
mechanisms: PointFlow~\citep{yang2019pointflow} uses continuous normalizing flows,
PointGrow~\citep{sun2020pointgrow} generates points autoregressively,
and ShapeGF~\citep{cai2020learning} learns spatial gradient fields.
In particular, PointFlow and ShapeGF represent shapes through distributions over
spatial locations and can sample a user-selected number of points.
This flexibility concerns the sampling resolution of a shape;
these models do not learn the distribution of cardinality or its
dependence on geometry.

\paragraph{Diffusion Models for Spatial Point Processes.}
For data with variable cardinality, some methods adopt a hierarchical
generation process. The equivariant diffusion model
(EDM)~\citep{hoogeboom2022equivariant}, referred to as the
fixed-dimensional diffusion model (FDDM) in our comparisons,
factorizes the point-set distribution as $p(x)=p(n)p(x\mid n)$:
it first samples cardinality, then generates all points jointly using a
cardinality-conditioned, permutation-equivariant denoiser.
\citet{mikuni2023fast} similarly use two diffusion models: one generates
the particle count and global properties of the particle collection,
such as momentum and mass, and the other jointly generates the particles'
features conditioned on them.
These hierarchical formulations can capture dependencies among points
and between cardinality and spatial configuration, but the sampled
cardinality remains fixed throughout point-level diffusion.

An alternative is to let cardinality and point locations evolve together
during generation.
Trans-dimensional diffusion (TDDM)~\citep{campbell2023trans} combines
continuous diffusion with discrete jumps. Its forward process diffuses
coordinates and deletes points according to a prescribed rate, stopping
deletions at a single point. The reverse process interleaves denoising
with insertions governed by a learned, state-dependent jump rate and an
insertion distribution. This design requires coordinating insertion with
denoising so that newly added points have enough time to be refined.
For molecules, the same process jointly diffuses coordinates and
continuous encodings of atom types and charges.

Another approach, Point Set Diffusion (PSD)~\citep{ludke2024unlocking},
evolves point sets through thinning and superposition.
Its forward process independently removes data points
and adds points from a noise process, eventually approaching that noise
process. In reverse, a classifier identifies retained data points, while
a conditional model predicts the number and locations of missing points,
using a learned Gaussian mixture for locations on Euclidean domains.
These predictions are combined with analytic thinning posteriors to
remove noise points and restore data points. Retained coordinates remain
fixed, so adjusting them requires deletion and resampling rather than
gradual coordinate updates. The published experiments do not demonstrate
joint generation of categorical point attributes.

Building on the goal of evolving cardinality and geometry together,
our EFDM represents point presence continuously and denoises it jointly
with coordinates and categorical attributes. This representation avoids
the explicit point-insertion and point-deletion steps used by TDDM and
PSD (Section~\ref{subsec:efdm}).

	\section{Preliminaries}
\subsection{Spatial Point Processes}
\label{sec:pre_spp}
Let $(D,d)$ be a complete separable metric space equipped with its Borel
$\sigma$-algebra $\mathcal{B}$.
A spatial point process (SPP)~\citep{gelfand2010handbook} is defined as a
random configuration of points in $D$.
Formally, let $(\Omega,\mathcal{F},\mathbb{P})$ be a probability space.
A spatial point process is a measurable mapping
\begin{equation*}
x : \Omega \rightarrow \mathcal{N}_{\text{lf}}(D),
\end{equation*}
where $\mathcal{N}_{\text{lf}}(D)$ denotes the set of locally finite point
configurations on $D$, i.e., subsets $x \subseteq D$ such that every bounded
Borel set contains only finitely many points.

Throughout this work, we additionally assume finite total cardinality
almost surely, $|x|<\infty$.
A realization can then be represented as a finite point set:
\begin{equation*}
x = \{x_1, \ldots, x_n\}, \quad x_i \in D,
\end{equation*}
where $n=|x|$ is the number of points.

\subsection{Diffusion Models}

Diffusion models are generative models that learn a data distribution
by gradually diffusing a data point with random noise until it reaches a prior distribution. They then learn a reverse process that can reconstruct the original data distribution via denoising. In discrete time, the forward process is defined as a Markov chain $q(x^{(1:T)} \mid x^{(0)}) := \prod_{t=1}^{T} q(x^{(t)} \mid x^{(t-1)}) $, where $q(x^{(t)} \mid x^{(t-1)})$ is the Markov diffusion kernel. The reverse process is defined as $p_{\theta}\!\left(x^{(0:T)} \right) := p\!\left(x^{(T)}\right)
\prod_{t=1}^{T} p_{\theta}\!\left(x^{(t-1)} \mid x^{(t)}\right)$, approximated by a trained neural network with parameters $\theta$.

	\section{Existence-Field Diffusion Model}
We first formulate the problem of modeling variable-cardinality point
sets, then introduce EFDM and its extension to categorical attributes.
\subsection{Problem Formulation}

Specializing the setting in Section~\ref{sec:pre_spp} to
$D=\mathbb{R}^d$, let
\begin{equation*}
\mathcal{X} = \{ x \subset \mathbb{R}^d : |x| < \infty \}
\end{equation*}
denote the configuration space. Given a dataset $\mathcal{D}$ of i.i.d. samples
from an unknown distribution $p^*(x)$ on $\mathcal{X}$, our goal is to
learn a diffusion-based model $p_\theta(x)$ that approximates $p^*(x)$
and generates new point sets. The model must satisfy two requirements.

\paragraph{Permutation invariance.}
When operating on ordered point representations, the model must assign
the same density to every ordering of a configuration. For any
permutation $\pi$ of $\{1,\dots,n\}$, this requires
$p_\theta(x_1,\dots,x_n) = p_\theta(x_{\pi(1)},\dots,x_{\pi(n)})$.

\paragraph{Variable cardinality.}
The model must learn the joint distribution of cardinality and spatial
configuration, capturing dependencies among points as well as between
their arrangement and point count.

\subsection{Existence-Field Point Diffusion}
\label{subsec:efdm}

We introduce the existence-field diffusion model (EFDM) for
variable-cardinality point sets. EFDM jointly diffuses point coordinates
and continuous existence variables, allowing spatial structure and
cardinality to be generated together without specifying the point count
in advance or relying on discrete point additions and deletions during
diffusion.
We first describe binary existence, then extend the representation to
points with categorical attributes.

\subsubsection{Binary Existence Fields}
\label{sec:efdm_binary}

For each potential point $x_i\in\mathbb{R}^d$, we introduce a real-valued
existence variable $u_i\in\mathbb{R}$. Its existence probability is
$e_i=\sigma(u_i)\in(0,1)$, where $\sigma$ is the sigmoid function.
The joint diffusion state is
\begin{equation}
 y=(y_1,\ldots,y_N)\in(\mathbb{R}^{d+1})^N,
 \qquad y_i=(x_i,u_i).
 \label{eq:efdm_binary_state}
\end{equation}

We represent each configuration using a fixed budget of $N$ potential
points, so EFDM models distributions on the bounded-cardinality space
\begin{equation}
 \mathcal{X}_N=\{x\subset\mathbb{R}^d:|x|\leq N\}\subset\mathcal{X}.
 \label{eq:efdm_bounded_space}
\end{equation}
In principle, $N$ can be chosen arbitrarily large. In practice, we choose
a finite slot budget to control the size of the diffusion state and
meet computational and memory budgets.
In the experiments, $N$ is chosen to be at least as large as the maximum
cardinality in the training data, so that no observed point set is truncated.

To represent a point set $x=\{x_1,\ldots,x_n\}$ with $n\leq N$, we assign
its points to $n$ active slots and use the remaining slots as auxiliary
padding. Active and inactive slots receive positive and negative existence
logits, respectively. The encoding and padding settings used in the
experiments are specified in Appendix~\ref{app:imp_detail}.

\paragraph{Joint diffusion.}
In the discrete-time formulation, we use the Gaussian forward process of
a denoising diffusion probabilistic model (DDPM)~\citep{ho2020denoising}
together with an existence-weighted denoising objective:
\begin{align}
 y^{(t)}
 &=\sqrt{\bar\alpha_t}\,y^{(0)}
   +\sqrt{1-\bar\alpha_t}\,\epsilon,
 \qquad \epsilon\sim\mathcal{N}(0,I),
 \label{eq:efdm_forward}\\
 \mathcal{L}(\theta)
 &=\mathbb{E}_{y^{(0)},t,\epsilon}
   \left[
   \frac{1}{Nd}\sum_{i=1}^{N}\sigma(u_i^{(t)})
   \left\|\epsilon_{x,i}-\hat\epsilon_{x,i}\right\|_2^2
   +\frac{1}{N}\sum_{i=1}^{N}
   \left(\epsilon_{u,i}-\hat\epsilon_{u,i}\right)^2
   \right],
 \label{eq:efdm_loss}
\end{align}
where $\bar\alpha_t=\prod_{s=1}^{t}(1-\beta_s)$,
$\epsilon_i=(\epsilon_{x,i},\epsilon_{u,i})$, and
$(\hat\epsilon_{x,i},\hat\epsilon_{u,i})=\epsilon_\theta(y^{(t)},t)_i$.
The coordinate denoising error is weighted by the existence probability
of each noisy slot, while the coordinate and existence losses are
averaged separately.
Here $y^{(0)}$ is sampled from the encoded training data distribution,
$t$ is uniform on $\{1,\ldots,T\}$, and the noise schedule
$\beta_s\in(0,1)$ is chosen so that the terminal forward distribution is
approximately standard Gaussian. In this discrete-time formulation,
sampling starts from $y^{(T)}\sim\mathcal{N}(0,I)$ and uses DDPM reverse
transitions parameterized by $\epsilon_\theta$.
The representation also supports continuous-time variance-preserving
(VP) diffusion~\citep{song2020score} within the same Gaussian diffusion
framework. The formulation and sampler used for each dataset are
specified in Appendix~\ref{app:imp_detail}.
A single denoiser processes the full state $y$, jointly modeling
spatial structure and existence. We require slot-permutation equivariance,
$\epsilon_\theta(P y,t)=P\epsilon_\theta(y,t)$ for any slot permutation $P$.
Together with the exchangeable Gaussian prior and identical decoding rules
for all slots, this yields a permutation-invariant point-set distribution.

After denoising, write the generated state as
$\hat y_i^{(0)}=(\hat x_i,\hat u_i)$. Bernoulli readout gives
\begin{equation}
 \hat z_i\sim\operatorname{Bernoulli}\!\left(\sigma(\hat u_i)\right),
 \qquad
 \hat x=\{\hat x_i:\hat z_i=1\},
 \qquad
 \hat n=\sum_{i=1}^{N}\hat z_i\leq N.
 \label{eq:efdm_decode}
\end{equation}

\subsubsection{Extension to Attributes}
\label{sec:typed}

For a marked point set, each active point carries $M\geq1$ categorical
attributes. Attribute $m\in\{1,\ldots,M\}$ has $K_m$ meaningful values,
with $a_i^{(m)}\in\{1,\ldots,K_m\}$ denoting its value for point $i$.
We retain one existence variable $u_i$ per slot, with
$e_i=\sigma(u_i)$, and let $\boldsymbol{\rho}_i^{(m)}$ describe the
conditional distribution of attribute $m$ given that the point exists.
On the simplex interior, $\rho_{ik}^{(m)}>0$ and
$\sum_{k=1}^{K_m}\rho_{ik}^{(m)}=1$.
Introducing category $0$ for non-existence gives the full categorical field
\begin{equation}
 \gamma_{i0}^{(m)}=1-e_i,
 \qquad
 \gamma_{ik}^{(m)}=e_i\rho_{ik}^{(m)},\quad k=1,\ldots,K_m.
 \label{eq:efdm_shared_existence}
\end{equation}
Thus $\boldsymbol{\gamma}_i^{(m)}=(\gamma_{i0}^{(m)},\ldots,\gamma_{iK_m}^{(m)})$
lies in the probability simplex over $K_m+1$ categories.
The same scalar existence probability $e_i$ is shared across all attributes
of slot $i$, so the slot is jointly present or absent for all attributes.

To obtain unconstrained attribute coordinates, we fix category $1$ as
the reference within each conditional distribution:
\begin{equation}
 v_{i1}^{(m)}=0,
 \qquad
 v_{ik}^{(m)}=\log\frac{\rho_{ik}^{(m)}}{\rho_{i1}^{(m)}},
 \quad k=2,\ldots,K_m.
 \label{eq:efdm_conditional_logratio}
\end{equation}
The free coordinates are
$\boldsymbol{v}_i^{(m)}=(v_{i2}^{(m)},\ldots,v_{iK_m}^{(m)})\in\mathbb{R}^{K_m-1}$,
which uniquely determine
$\boldsymbol{\rho}_i^{(m)}=\operatorname{softmax}(0,\boldsymbol{v}_i^{(m)})$.
The joint diffusion state is
\begin{equation}
 y_i=(x_i,u_i,\boldsymbol{v}_i^{(1)},\ldots,\boldsymbol{v}_i^{(M)})
 \in\mathbb{R}^{d+1+\sum_{m=1}^{M}(K_m-1)}.
 \label{eq:efdm_attribute_state}
\end{equation}
We apply the Gaussian forward process in Eq.~\eqref{eq:efdm_forward}
to this state and extend the denoising objective to include errors on
the free attribute coordinates. A single permutation-equivariant denoiser
processes all slots, jointly modeling coordinates, existence, and
attributes. With no attributes, the state reduces to binary EFDM.
The experimental encoding of categorical labels is described in
Appendix~\ref{app:imp_detail}.

After denoising, we draw $\hat z_i$ once per slot as in
Eq.~\eqref{eq:efdm_decode}. For each active slot $\hat z_i=1$, we draw
$\hat a_i^{(m)}\sim\operatorname{Categorical}(\hat{\boldsymbol{\rho}}_i^{(m)})$ and return
$\hat x=\{(\hat x_i,\hat a_i^{(1)},\ldots,\hat a_i^{(M)}):\hat z_i=1\}$.
Inactive slots and all their attributes are discarded.
The categorical draws are independent conditional on the generated
fields, while the joint denoiser can model dependencies among the fields.
Alternatively, the generated categorical fields can be decoded
deterministically; the parameterization and readout used for QM9 are
specified in Section~\ref{sec:molecule}.

	\newcommand\tddmv{TDDMv}
\newcommand{\rowlabel}[1]{%
	\raisebox{2.8em}[0pt][0pt]{\rotatebox{90}{\textbf{#1}}}%
}
\section{Experiments}
\label{sec:experiments}
We evaluate EFDM on molecular generation and spatial point-set modeling.
The main experiments cover the QM9 molecule
dataset~\citep{ruddigkeit2012enumeration,ramakrishnan2014quantum},
the Japan Earthquakes dataset~\citep{usgs2024earthquakes},
and the City of Louisville e-scooter trip dataset~\citep{lvscooter}.
We also use a controlled bent Gaussian-mixture synthetic benchmark to
study count--configuration dependence (Section~\ref{sec:synthetic_bent}).

\subsection{Baselines}
\label{sec:baselines}
We consider five baselines: FDDM, TDDM, PSD, Independent Point Diffusion
(IDM), and cardinality-adaptive TDDM (\tddmv{}).
FDDM, TDDM, and PSD are described in Section~\ref{sec:related_work}.

IDM samples cardinality separately and models points independently via
$p(x\mid n)\approx\prod_{i=1}^{n}p_{\mathrm{pt}}(x_i)$, where
$p_{\mathrm{pt}}$ is a pointwise diffusion model shared across cardinalities.

Following the sample-independent rate design of~\citet{campbell2023trans},
our base TDDM uses a shared deletion rate calibrated using the maximum
training cardinality minus one. Forward deletion counts are capped to
retain at least one point.
QM9 contains at most 29 atoms per molecule, including
hydrogens~\citep{hoogeboom2022equivariant}.
To better accommodate the larger point sets in the Earthquake and Trip
datasets, we introduce \tddmv{}, a variant that adapts the deletion rate to
each sample's initial cardinality.
It uses the observed initial count during training and the model's current
prediction of that initial count during generation.
Dataset-specific configurations are provided in Appendix~\ref{app:imp_detail}.

\subsection{Evaluation Metrics}
\label{sec:evaluation_metrics}
For QM9, we report atom stability, molecule stability, and validity
(Section~\ref{sec:molecule}). For spatial point sets, we assess
cardinality, spatial statistics, and their combined distributional behavior.
Count $W_1$ measures cardinality accuracy by comparing the distributions
of the number of points per set. Mean, Var, Skew, Kurt, and NN1 assess
spatial statistics: location, spread, asymmetry, excess kurtosis, and
local spacing, respectively. We report $W_1$ distances between the
generated and held-out distributions of these per-set summaries, with
coordinate-wise discrepancies averaged over dimensions.
MMD, JMMD$^2$, and ICMMD$^2$ complement these separate summaries with
distribution-level comparisons: MMD compares whole point sets, while
JMMD$^2$ and ICMMD$^2$ explicitly relate cardinality to spatial
configuration through joint and conditional comparisons, respectively.

We adopt maximum mean discrepancy (MMD) from~\citet{gretton2012kernel}
and reuse the implementation in PSD~\citep{ludke2024unlocking}, which
constructs a point-set kernel using Wasserstein distance (WD),
implemented with 50-projection sliced $W_2$.
PSD estimates its bandwidth from pairwise distances involving real test
and generated sets. To complement this evaluation-dependent choice, we
report $\mathrm{MMD}_{\mathrm{train\mbox{-}scale}}$, whose kernel scale is
fixed from training data and shared across methods; its scale convention
is specified in Appendix~\ref{app:metric_details}.

However, this Wasserstein-based MMD uses normalized point measures
without an explicit count term and does not separately assess spatial
agreement at each cardinality. We therefore use
JMMD$^2$~\citep{long2017joint} to compare the joint distribution of
cardinality and spatial configuration using an exact-count product kernel.
ICMMD$^2$, adapted from~\citet{yan2024conditional}, compares spatial
distributions conditional on exact cardinality, weighted by eligible real
count frequencies. It additionally penalizes insufficient generated
support. These two metrics report squared scores. Full definitions,
preprocessing, bandwidths,
estimators, and coverage diagnostics are given in
Appendix~\ref{app:metric_details}; dataset-specific evaluation diagnostics
are given in Appendix~\ref{app:spatial_evaluation}.

The Earthquake and Trip tables report the same ten metrics:
Mean, Var, Skew, Kurt, NN1, Count $W_1$, MMD, train-scale MMD, JMMD$^2$,
and ICMMD$^2$. The bent synthetic table in
Appendix~\ref{subsec:syn_bent_data_gen} reports the same ten metrics.
We retain unnormalized Count $W_1$ throughout. JMMD$^2$ is
displayed after multiplication by $10^3$. Across result tables,
bold marks the best reported value (the mean for repeated evaluations),
including ties at the displayed precision. Rankings are descriptive and
do not indicate statistical significance. We do not rank the
near-saturated Trip MMD.

\subsection{Experimental Results}
\label{sec:experimental_results}

\subsubsection{Molecular Generation on QM9}
\label{sec:molecule}
We evaluate unconditional molecular generation on QM9, which contains
small organic molecules with varying numbers of atoms.
Each molecule is represented as a set of atoms with 3D coordinates and
categorical atom types.
For QM9, we jointly diffuse atom coordinates and direct categorical
logits relative to a fixed zero logit for the inactive category.
Training uses $+1$ for the observed atom type and $-1$ for all other
type logits, with all type logits $-1$ for inactive slots. Since atom
type is the only attribute, a deterministic argmax over the type and
inactive logits jointly determines presence and type without additional
sampling;
Appendix~\ref{app:attribute_parameterization} details the parameterization
and readout.
We use continuous-time VP diffusion with predictor--corrector sampling
and an E(n)-equivariant graph neural network (EGNN)
backbone~\citep{satorras2021en} together with a Transformer
component~\citep{vaswani2017attention};
implementation details are given in Appendix~\ref{app:imp_detail}.

We compare EFDM with the published FDDM and TDDM scores from Table~2
of~\citet{campbell2023trans}.
The FDDM baseline corresponds to the equivariant diffusion model (EDM)
of~\citet{hoogeboom2022equivariant}.
We report the same three molecular quality metrics: atom stability,
molecule stability, and validity.
Bonds are inferred from interatomic distances.
Atom stability measures the fraction of atoms with the correct valency,
and molecule stability measures the fraction of molecules for which all
atoms are stable; validity is assessed using \citet{rdkit}.

Using the final exponential moving average (EMA) checkpoint at epoch
8,000, EFDM achieves 99.04\% atom stability, 91.18\% molecule stability,
and 96.38\% validity, exceeding the reported FDDM and TDDM scores on all
three metrics (Table~\ref{tab:qm9_results}). Compared with TDDM,
molecule stability and validity improve by 4.0 and 4.1 percentage
points, respectively.

\begin{table}[t]
	\centering
	\small
	\caption{
		Unconditional molecular generation on QM9.
		All metrics are percentages ($\uparrow$), evaluated on 10,000
		generated molecules.
		FDDM and TDDM scores are taken from Table~2
		of~\citet{campbell2023trans}; the FDDM scores originate
		from~\citet{hoogeboom2022equivariant}.
		EFDM is evaluated once using the final-epoch
		EMA checkpoint. Bold marks the
		highest reported value at the displayed precision, including ties.
	}
	\label{tab:qm9_results}
	\begin{tabular}{lccc}
		\toprule
		Method & Atom stability & Molecule stability & Validity \\
		\midrule
		EFDM & \textbf{99.04} & \textbf{91.18} & \textbf{96.38} \\
		FDDM & 98.7 & 82.0 & 91.9 \\
		TDDM & 98.3 & 87.2 & 92.3 \\
		\bottomrule
	\end{tabular}
\end{table}

\subsubsection{Earthquake Data}
\label{sec:earthquake}
We use the Japan Earthquakes dataset from the earthquake catalog
maintained by the \citet{usgs2024earthquakes}, following the dataset used by
Point Set Diffusion (PSD)~\citep{ludke2024unlocking}.
Each observation is a spatial point set of earthquake locations within a
week. The 1,500 weekly sets are randomly divided into 900 training,
300 validation, and 300 test sets for each of three split/training seeds.
This evaluates generation of weekly spatial patterns rather than
forecasting future weeks.

We compare EFDM with FDDM, TDDM,
\tddmv{}, IDM, and PSD.
FDDM and IDM sample cardinalities from a distribution fitted to the
training counts before coordinate generation; EFDM and the
trans-dimensional baselines determine cardinality within the generative
process.
All methods are trained for 60,000 epochs, and each checkpoint is selected
by the lowest EMA validation loss.
We sample without a corrector, rejecting empty proposals and resampling
until 4,096 nonempty point sets are obtained per method and seed.
All ten metrics, including count $W_1$, use these accepted sets against
all 300 test sets.
Implementation and evaluation details are given in
Appendix~\ref{app:imp_detail}.

%Table~\ref{tab:eq_results} shows that EFDM has lower mean errors than PSD in all ten reported metrics.
%Overall, EFDM achieves state-of-the-art performance in capturing complex spatial structures and conditional spatial-cardinality dependencies, attaining the best scores in the mean train-scale MMD, ICMMD, and three out of five spatial summary metrics (Mean, Skewness, and Kurtosis). 
%While two-stage factorized baselines such as FDDM benefit from separately fitting the marginal cardinality distribution $p(n)$ before conditional coordinate generation, EFDM achieves highly competitive joint alignment (JMMD$^2 = 0.407 \times 10^{-3}$) through unified end-to-end diffusion. These results support the effectiveness of continuous existence fields in jointly capturing cardinality and spatial structure without a separate cardinality model.

Table~\ref{tab:eq_results} shows that EFDM achieves lower mean errors than PSD across all ten reported metrics and strong overall performance among the evaluated methods. EFDM obtains the best mean scores for train-scale MMD, ICMMD, and three of the five spatial-summary metrics (Mean, Skewness, and Kurtosis).
EFDM is consistently second-best in variance, NN1, MMD, and JMMD, with FDDM leading these metrics through a two-stage formulation that separately fits the marginal cardinality distribution $p(n)$ before conditional coordinate generation.
For Count $W_1$, FDDM and IDM use the same separately fitted cardinality model to sample $n$ before coordinate generation, whereas EFDM generates cardinality and spatial structure jointly within a single diffusion process. Despite not using a separate cardinality model, EFDM achieves a Count $W_1$ of $6.2$ and a $\mathrm{JMMD}^2$ of $0.407\times 10^{-3}$. These results support continuous existence fields as an effective approach to joint cardinality--spatial modeling.

%Across all six methods, EFDM has the lowest mean train-scale MMD and ICMMD scores and the lowest mean, skewness, and kurtosis summary errors; FDDM has the lowest count error, variance and NN1 errors, MMD, and JMMD.
%Thus EFDM's shared diffusion process gives competitive spatial fidelity, while FDDM's factorized model also provides a strong joint fit.

\begin{table}[t]
 \centering
 \footnotesize
 \setlength{\tabcolsep}{3.5pt}
 \caption{Japan Earthquakes. All ten metrics use the same 4,096 accepted
 nonempty generated sets per run against 300 test sets.
 Values are mean $\pm$ sample standard deviation
 across three split/training seeds, computed directly from the unrounded
 per-seed metrics.
 All errors are lower-is-better; bold marks the lowest mean at the
 displayed precision, including ties.}
 \label{tab:eq_results}
 \resizebox{\linewidth}{!}{%
 \begin{tabular}{lccccc}
  \toprule
  Method & Mean & Var & Skew & Kurt & NN1 \\
  \midrule
  EFDM & $\mathbf{0.0271\pm0.0128}$ & $0.0295\pm0.0116$ & $\mathbf{0.0547\pm0.0130}$ & $\mathbf{0.1869\pm0.0110}$ & $0.0097\pm0.0022$ \\
  FDDM & $0.0290\pm0.0155$ & $\mathbf{0.0291\pm0.0094}$ & $0.0582\pm0.0143$ & $0.1910\pm0.0369$ & $\mathbf{0.0045\pm0.0011}$ \\
  TDDM & $0.1993\pm0.0136$ & $0.8057\pm0.0177$ & $1.5626\pm0.2349$ & $13.3865\pm0.4090$ & $0.0932\pm0.0046$ \\
  \tddmv{} & $0.0441\pm0.0174$ & $0.0374\pm0.0080$ & $0.1374\pm0.0217$ & $0.4117\pm0.0653$ & $0.0354\pm0.0026$ \\
  IDM & $0.1044\pm0.0155$ & $0.0827\pm0.0069$ & $0.2348\pm0.0165$ & $0.6124\pm0.0631$ & $0.0212\pm0.0035$ \\
  PSD & $0.0563\pm0.0119$ & $0.0450\pm0.0094$ & $0.1305\pm0.0136$ & $0.3537\pm0.0608$ & $0.0452\pm0.0068$ \\
  \bottomrule
 \end{tabular}
 }
 \par\vspace{0.5em}
 \resizebox{\linewidth}{!}{%
 \begin{tabular}{lccccc}
  \toprule
  Method & Count $W_1$ & MMD & $\mathrm{MMD}_{\mathrm{train\mbox{-}scale}}$ & JMMD$^2$ ($\times10^3$) & ICMMD$^2$ \\
  \midrule
  EFDM & $6.2\pm1.5$ & $0.0701\pm0.0118$ & $\mathbf{0.0498\pm0.0142}$ & $0.407\pm0.197$ & $\mathbf{0.0085\pm0.0105}$ \\
  FDDM & $\mathbf{4.1\pm0.3}$ & $\mathbf{0.0679\pm0.0105}$ & $0.0500\pm0.0135$ & $\mathbf{0.265\pm0.224}$ & $0.0103\pm0.0069$ \\
  TDDM & $224.5\pm20.2$ & $0.0729\pm0.0058$ & $0.8078\pm0.0105$ & $23.821\pm4.018$ & $1.5419\pm0.0271$ \\
  \tddmv{} & $27.9\pm2.9$ & $0.0798\pm0.0120$ & $0.0751\pm0.0176$ & $5.732\pm0.726$ & $0.3130\pm0.0149$ \\
  IDM & $4.7\pm0.6$ & $0.1249\pm0.0052$ & $0.1752\pm0.0133$ & $0.694\pm0.184$ & $0.0579\pm0.0111$ \\
  PSD & $18.3\pm5.7$ & $0.0922\pm0.0095$ & $0.0924\pm0.0111$ & $1.890\pm0.360$ & $0.0700\pm0.0288$ \\
  \bottomrule
 \end{tabular}
 }
\end{table}

\subsubsection{Trip Data}
\label{sec:trip}
We use the City of Louisville e-scooter trip dataset, which contains
real-world shared micro-mobility trip records in Louisville, Kentucky.
Each sample consists of approximately 10\% of the pick-up locations for
a given day, after filtering using statistics fitted on the training data.
For each split/training seed, the days are randomly divided into
268 training, 89 validation, and 91 test sets.

We compare EFDM with FDDM, TDDM, \tddmv{}, IDM, and PSD, with the
method-specific configurations detailed in Appendix~\ref{app:imp_detail}.
All methods are trained for 60,000 epochs and evaluated at the checkpoint
with the lowest EMA validation loss.
We sample without a corrector, rejecting empty proposals and resampling
until 4,096 nonempty point sets are obtained per method and seed.
All ten metrics, including count $W_1$, use these accepted sets against
the full 91-set test split. We report mean $\pm$ sample
standard deviation over the three seeds.

%EFDM has the lowest mean train-scale MMD, JMMD$^2$, ICMMD$^2$, and four spatial-summary errors (Table~\ref{tab:trip_feature_w1}). FDDM, IDM, and PSD tie on count at the displayed precision; FDDM leads on kurtosis.
%EFDM has a higher mean count error than FDDM, IDM, and PSD, while its JMMD$^2$ estimates show substantial variation across seeds.
EFDM achieves the lowest mean train-scale MMD, JMMD$^2$, and ICMMD$^2$ scores among the compared methods (Table~\ref{tab:trip_feature_w1}). It also achieves the lowest mean errors on four of the five spatial-summary metrics and ranks second on kurtosis.

Evaluation details are in Appendix~\ref{app:spatial_evaluation}.
Coordinate normalizations and test-set sizes differ, so absolute errors
are compared only within each dataset.

\begin{table}[t]
 \centering
 \footnotesize
 \setlength{\tabcolsep}{3.5pt}
 \caption{Louisville Trip: all metrics use the same 4,096 accepted
 nonempty generated sets and 91 test sets.
 Values are mean $\pm$ sample standard deviation over three seeds;
 lower is better. Bold marks the lowest displayed mean, including ties.
 The original MMD is nearly constant across methods and is therefore
 not ranked. The unbiased JMMD$^2$ estimate may be negative.}
 \label{tab:trip_feature_w1}
 \resizebox{\linewidth}{!}{%
 \begin{tabular}{lccccc}
  \toprule
  Method & Mean & Var & Skew & Kurt & NN1 \\
  \midrule
  EFDM & $\mathbf{0.0576\pm0.0090}$ & $\mathbf{0.0602\pm0.0103}$ & $\mathbf{0.1138\pm0.0437}$ & $0.3805\pm0.0201$ & $\mathbf{0.0151\pm0.0119}$ \\
  FDDM & $0.0635\pm0.0129$ & $0.0842\pm0.0306$ & $0.1150\pm0.0490$ & $\mathbf{0.3697\pm0.0663}$ & $0.0159\pm0.0083$ \\
  TDDM & $0.3348\pm0.0055$ & $0.2323\pm0.0339$ & $0.5648\pm0.0661$ & $1.9065\pm0.1142$ & $0.0632\pm0.0104$ \\
  \tddmv{} & $0.1024\pm0.0254$ & $0.1050\pm0.0088$ & $0.2062\pm0.0416$ & $0.6367\pm0.2265$ & $0.0456\pm0.0070$ \\
  IDM & $0.2042\pm0.0132$ & $0.1467\pm0.0103$ & $0.3057\pm0.0279$ & $0.7087\pm0.1727$ & $0.0304\pm0.0085$ \\
  PSD & $0.0647\pm0.0104$ & $0.0787\pm0.0161$ & $0.1295\pm0.0307$ & $0.3976\pm0.1282$ & $0.0343\pm0.0082$ \\
  \bottomrule
 \end{tabular}
 }
 \par\vspace{0.5em}
 \resizebox{\linewidth}{!}{%
 \begin{tabular}{lccccc}
  \toprule
  Method & Count $W_1$ & MMD & $\mathrm{MMD}_{\mathrm{train\mbox{-}scale}}$ & JMMD$^2$ ($\times10^3$) & ICMMD$^2$ \\
  \midrule
  EFDM & $13.81\pm7.36$ & $0.106005\pm2.04\!\times\!10^{-6}$ & $\mathbf{0.0756\pm0.0033}$ & $\mathbf{-0.151\pm1.452}$ & $\mathbf{0.0123\pm0.0085}$ \\
  FDDM & $\mathbf{10.93\pm2.92}$ & $0.105997\pm4.67\!\times\!10^{-6}$ & $0.0782\pm0.0041$ & $0.145\pm1.515$ & $0.0195\pm0.0236$ \\
  TDDM & $141.99\pm10.55$ & $0.105987\pm1.11\!\times\!10^{-11}$ & $0.4050\pm0.0149$ & $23.296\pm1.319$ & $1.5403\pm0.0807$ \\
  \tddmv{} & $50.81\pm8.12$ & $0.106130\pm5.25\!\times\!10^{-5}$ & $0.1292\pm0.0126$ & $5.921\pm2.890$ & $0.2229\pm0.2112$ \\
  IDM & $\mathbf{10.93\pm2.92}$ & $0.105987\pm1.83\!\times\!10^{-7}$ & $0.2427\pm0.0064$ & $0.802\pm1.619$ & $0.1191\pm0.0172$ \\
  PSD & $\mathbf{10.93\pm6.14}$ & $0.106001\pm6.14\!\times\!10^{-6}$ & $0.0866\pm0.0049$ & $0.240\pm1.718$ & $0.0126\pm0.0192$ \\
  \bottomrule
 \end{tabular}
 }
\end{table}

\subsubsection{Synthetic Data: Bent Gaussian Mixtures}
\label{sec:synthetic_bent}
In this two-component Gaussian benchmark, a latent variable couples
cardinality and component separation. The fixed pre-bend width is
$\sigma_{\mathrm{data}}=0.2$; the invertible bend
$(x,y)\mapsto(x,y+x^2)$ preserves their dependence.

% Reproducible vector figure from archived summaries and frozen trends.
% Build with figures/synthetic_bent/make_figure2.py.
\begin{figure}[!htbp]
 \centering
 \includegraphics[width=0.80\linewidth]{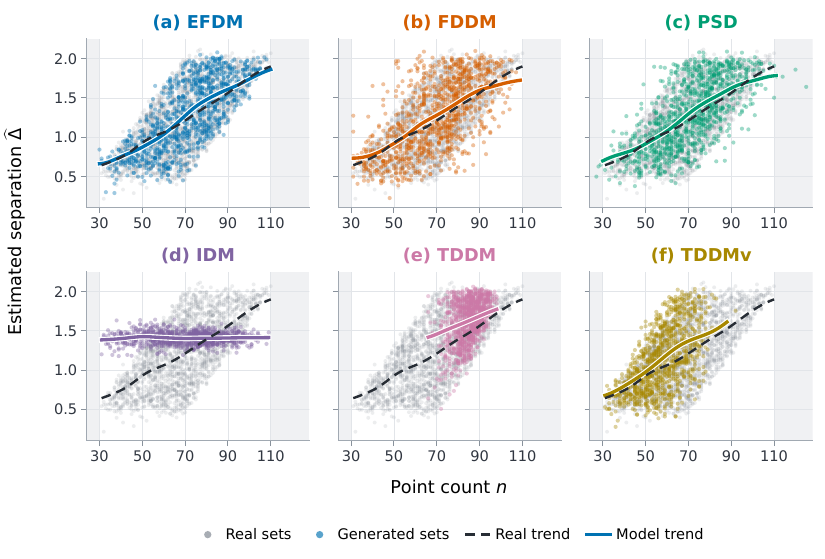}
 \caption{Bent synthetic count versus estimated separation. Each panel
 compares 1,024 generated sets (colored) with the same 2,000 real test
 sets (gray); solid/dashed curves show their trends. Shading marks counts
 outside $[30,110]$. One training seed; estimator and smoothing details
 are in Appendix~\ref{subsec:syn_bent_data_gen}.}
 \label{fig:syn_bent_count_spacing}
\end{figure}

Figure~\ref{fig:syn_bent_count_spacing} compares count and estimated
separation after the same inverse bend. At larger counts, EFDM tracks the
real-test trend more closely than FDDM and PSD, whose curves flatten;
IDM is nearly count-independent, TDDM covers a narrow count range, and
\tddmv{} favors lower counts.

EFDM achieves the lowest JMMD$^2$ and full ICMMD$^2$ scores, while FDDM and PSD lead on several spatial summaries (Table~\ref{tab:syn_bent_results}). 
Full data and evaluation details are provided in Appendix~\ref{subsec:syn_bent_data_gen}.

	% Keep this short section together so preceding floats cannot split it.
\Needspace{11\baselineskip}
\begin{samepage}
\section{Conclusions}
\label{sec:conclusion}
EFDM brings variable-cardinality point sets into a unified Gaussian diffusion framework, allowing point presence, geometry, and categorical attributes to be refined jointly without fixing cardinality before denoising or introducing explicit birth–death transitions. Its improved stability and validity on the molecular dataset and competitive performance on real-world and synthetic spatial datasets demonstrate the potential of continuous existence fields for point-set generation. In practice, the chosen slot budget sets the maximum supported cardinality, and larger budgets involve additional computation and memory.
\par
\end{samepage}

	\clearpage
	\subsection*{AI Use Statement}
AI tools were only used to assist with manuscript editing and polishing,
code implementation and debugging for experiments and synthetic data
generation, and the presentation of mathematical derivations.
All AI-assisted content was reviewed and verified by the authors.
The authors take full responsibility for the content of this work.

\subsection*{Reproducibility Statement}
Details required to reproduce the experiments are provided in the appendices to this paper. Appendix~\ref{subsec:syn_bent_data_gen} specifies the synthetic-data construction and evaluation protocol; Appendices~\ref{app:imp_detail}--\ref{app:attribute_parameterization} describe model configurations, training, sampling, and categorical parameterizations; and Appendices~\ref{app:metric_details}--\ref{app:spatial_evaluation} provide the complete metric definitions and spatial evaluation protocols.
Source code for training, sampling, and evaluation is provided in the supplementary code package.

\subsection*{Ethics Statement}
This work develops generative modeling methodology using synthetic and publicly available scientific and spatial datasets. It does not involve human-subject experiments or collection of private personal data.

	\bibliographystyle{plainnat}
	\bibliography{bib/references}
	
	\appendix
\section{Supplementary Material}
\label{app:sup_mat}

This appendix provides bent synthetic data generation details and complete experimental results, model configurations, QM9 evaluation, and spatial evaluation metrics and protocols, together with supplementary visualizations for the QM9, Earthquake, and Trip experiments.

\subsection{Bent Gaussian Mixtures: Data and Results}
\label{subsec:syn_bent_data_gen}

This section gives the data-generation rules, evaluation protocol, and
complete results for the bent Gaussian-mixture benchmark introduced in
Section~\ref{sec:synthetic_bent}, together with estimation and smoothing
details for
Figure~\ref{fig:syn_bent_count_spacing}.

\paragraph{Base distribution.}
Each point set has two independent latent variables
$a,b\overset{\mathrm{iid}}{\sim}\mathrm{Unif}(0,1)$. They determine
its cardinality and component separation as
\begin{equation}
 n=\operatorname{round}\!\left(30+40(a+b)\right),
 \qquad \Delta=0.5+1.5a,
 \qquad \sigma_{\mathrm{data}}=0.2.
 \label{eq:bent_latents}
\end{equation}
Thus $30\leq n\leq110$, and cardinality is statistically related to
separation without determining it uniquely. Conditional on these
latents, the unbent points are sampled independently as
\begin{equation}
 \boldsymbol{\xi}_i\mid a,b\overset{\mathrm{iid}}{\sim}
 \tfrac12\mathcal{N}\!\left(
   \begin{pmatrix}-\Delta/2\\0\end{pmatrix},
   \sigma_{\mathrm{data}}^2 I_2\right)
 +\tfrac12\mathcal{N}\!\left(
   \begin{pmatrix}\Delta/2\\0\end{pmatrix},
   \sigma_{\mathrm{data}}^2 I_2\right),
 \qquad i=1,\ldots,n.
 \label{eq:bent_base_mixture}
\end{equation}
Component assignments are independent Bernoulli draws: the two
components have equal sampling probability, but are not forced to
contain equal numbers of points in an individual set. We randomly
permute the sampled points before storing the set.

\paragraph{Nonlinear deformation and pairing.}
The observed points are obtained through the pointwise shear
\begin{equation}
 \boldsymbol{x}_i=T_\kappa(\boldsymbol{\xi}_i),
 \qquad
 T_\kappa(r,s)=(r,s+\kappa r^2),
 \qquad \kappa=1.
 \label{eq:bent_transform}
\end{equation}
Its inverse is $T_\kappa^{-1}(x,y)=(x,y-\kappa x^2)$, and its
Jacobian determinant is one. The transformation changes the geometry
while retaining the information in the original point set and its
cardinality. It does not introduce additional interactions between
points: conditional independence given the shared latents is
preserved. The fixed $\sigma_{\mathrm{data}}$ describes the Gaussian
width before bending, not a constant local Euclidean width after
the nonlinear transformation.

The bent and unbent datasets are paired point by point. They retain
the same latent draws, cardinalities, component labels, point order,
and train/validation/test membership. We use 500 training, 200
validation, and 2,000 test sets. The latent variables and generating
parameters are retained as diagnostic metadata, not additional model
inputs. The transformation is evaluated
in float64 before converting the coordinates to float32; existing
$+\infty$ padding is retained and is not transformed. No centering,
rescaling, or clipping is applied to construct the bent dataset.
The bounds used by the evaluation adapter are recomputed from the
bent training split.

\paragraph{Training and evaluation.}
All six methods use the same 500 training, 200 validation, and 2,000 test
sets.
The six reported models are trained from scratch for 60,000 epochs
with Adam~\citep{kingma2015adam} and training seed 2025. The set-based
models use four-layer networks with 128-dimensional embeddings,
256-dimensional feed-forward widths, and four attention heads; IDM uses
a four-layer point-wise MLP without attention. Training batches contain
64 point sets for EFDM, FDDM, TDDM, \tddmv{}, and PSD, whereas IDM uses
4,096 individual points per batch. The learning rate is $10^{-4}$ with
zero weight decay except for PSD, which uses learning rate $10^{-3}$,
weight decay $10^{-4}$, and gradient clipping at 2.
EMA decays by 0.999 after each update. Every 100 epochs, we evaluate the
EMA model with fixed validation seed 123 and select the lowest-loss
checkpoint among 600 candidates for each method. Each selected model
generates 1,024 point sets with sampling seed 42, which are compared
against all 2,000 test sets. Non-PSD models use 1,000 predictor steps
without a corrector; PSD uses its native 100-step sampler. These are
single-run results, not means over training seeds. All 1,024 generated
sets for each method are nonempty.

Feature $W_1$ is evaluated in the observed bent coordinates, using
all generated sets, including any sets with cardinality
outside $[30,110]$.
Cardinality $W_1$ uses every generated set. Structural spacing and
width diagnostics first apply the same inverse map
$T_1^{-1}(x,y)=(x,y-x^2)$ to both real and generated points.
For JMMD, ICMMD, and the spatial MMD variants, the evaluation adapter
maps coordinates to a bounded domain and clips values outside its
training-derived bounds. This metric preprocessing is distinct from
the unclipped data-generation process.

JMMD and ICMMD use the shared training-derived baseline bandwidth
$\sigma=0.133825$ and the definition in
Appendix~\ref{app:metric_details}. Train-scale MMD uses
$h=0.161057$, the median distance over all ordered training pairs
including self-pairs, with kernel $\exp[-d_{\mathrm{SW}}/(2h)]$.
Here $d_{\mathrm{SW}}$ denotes the unsquared sliced-Wasserstein
distance. The bandwidths are shared across
methods; none is fitted to a method's test performance.

\paragraph{Quantitative results.}
Table~\ref{tab:syn_bent_results} shows that EFDM achieves the lowest
JMMD$^2$ and full ICMMD$^2$ scores. Its measured joint discrepancy is 68.3\% lower
than FDDM's and 55.9\% lower than PSD's.
FDDM has the lowest cardinality error and both MMD variants, as well as
the lowest mean, variance, and kurtosis summary errors; PSD has the
lowest skewness and NN1 errors. These results illustrate that strong
count and spatial-summary scores can coexist with a larger discrepancy
under the joint metric.
EFDM retains the lowest JMMD$^2$ at half and twice the
training bandwidth (Table~\ref{tab:bent_bandwidth}).

\begin{table}[!htbp]
 \centering
 \footnotesize
 \setlength{\tabcolsep}{3.5pt}
 \caption{Bent Gaussian mixtures with fixed pre-transform
 $\sigma_{\mathrm{data}}=0.2$ and $\kappa=1$. Results use a single training
 seed and 1,024 generated sets against 2,000 test sets. The upper panel
 reports spatial-summary $W_1$ in the observed bent coordinates. All errors
 are lower-is-better. Bold marks the lowest value at the displayed
 precision, including ties.}
 \label{tab:syn_bent_results}
 \begin{tabular*}{\linewidth}{@{\extracolsep{\fill}}lccccc@{}}
  \toprule
  Method & Mean & Var & Skew & Kurt & NN1 \\
  \midrule
  EFDM & $0.02250$ & $0.02214$ & $0.02432$ & $0.04479$ & $0.001179$ \\
  FDDM & $\mathbf{0.01805}$ & $\mathbf{0.01742}$ & $0.01627$ & $\mathbf{0.03650}$ & $0.000770$ \\
  TDDM & $0.13729$ & $0.15160$ & $0.03881$ & $0.15748$ & $0.006640$ \\
  \tddmv{} & $0.03388$ & $0.03440$ & $0.03839$ & $0.07866$ & $0.005797$ \\
  IDM & $0.10742$ & $0.14532$ & $0.12836$ & $0.13562$ & $0.016445$ \\
  PSD & $0.02688$ & $0.02694$ & $\mathbf{0.01451}$ & $0.04262$ & $\mathbf{0.000699}$ \\
  \bottomrule
 \end{tabular*}
 \par\vspace{0.5em}
 \begin{tabular*}{\linewidth}{@{\extracolsep{\fill}}lccccc@{}}
  \toprule
  Method & Count $W_1$ & MMD & $\mathrm{MMD}_{\mathrm{train\mbox{-}scale}}$ & JMMD$^2$ ($\times10^3$) & ICMMD$^2$ \\
  \midrule
  EFDM & $2.078$ & $0.0542$ & $0.0554$ & $\mathbf{0.167}$ & $\mathbf{0.0249}$ \\
  FDDM & $\mathbf{1.222}$ & $\mathbf{0.0448}$ & $\mathbf{0.0426}$ & $0.527$ & $0.0292$ \\
  TDDM & $16.391$ & $0.2574$ & $0.3537$ & $22.110$ & $0.8821$ \\
  \tddmv{} & $11.677$ & $0.0665$ & $0.0844$ & $5.188$ & $0.4307$ \\
  IDM & $1.470$ & $0.3305$ & $0.3582$ & $5.426$ & $0.3750$ \\
  PSD & $2.052$ & $0.0583$ & $0.0640$ & $0.379$ & $0.0297$ \\
  \bottomrule
 \end{tabular*}
\end{table}

\paragraph{Count--separation visualization.}
For Figure~\ref{fig:syn_bent_count_spacing}, let $r_i=y_i-x_i^2$ be the
vertical coordinate after the inverse transformation. We use the same
uncentered moment estimator for real and generated sets:
\begin{equation}
 \widehat{\Delta}(S)=2\sqrt{\max\!\left\{
 \frac{1}{n}\sum_{i=1}^{n}x_i^2-
 \frac{1}{n}\sum_{i=1}^{n}r_i^2,\,0\right\}}.
 \label{eq:bent_spacing_estimator}
\end{equation}
The plotted quantity is an estimate from observed coordinates, not the
latent component separation. Trend curves average $\widehat{\Delta}$ with Gaussian
weights $w_i(n)=\exp[-(n_i-n)^2/(2h^2)]$, where $n_i$ is the cardinality
of set $i$ and $h=5$ count units for every method.
Curves are shown only within each collection's observed
count range and where $(\sum_iw_i)^2/\sum_iw_i^2\ge20$. All scatter points
are retained, including the eight PSD samples outside $[30,110]$.
The same real-test samples and trend are used in all panels.

\paragraph{Spatial summaries and bandwidth sensitivity.}
The upper panel of Table~\ref{tab:syn_bent_results} reports the five
spatial-summary errors. Table~\ref{tab:bent_bandwidth} varies the JMMD
bandwidth to $0.5\sigma$, $\sigma$, and $2\sigma$, leaving the
trained models and generated sets unchanged. EFDM has the lowest
JMMD$^2$ in all three settings.

\begin{table}[t]
\centering
\small
\caption{Bandwidth sensitivity for bent two-Gaussian data. The
columns vary the common training-derived bandwidth
$\sigma=0.133825$. JMMD$^2$ is displayed after multiplication by $10^3$.
All scores are lower-is-better.
Bold marks the lowest value at the displayed precision, including ties.}
\label{tab:bent_bandwidth}
\setlength{\tabcolsep}{7pt}
\begin{tabular}{lccc}
\toprule
Method & $0.5\sigma$ & $\sigma$ & $2\sigma$ \\
\midrule
EFDM & $\mathbf{0.133}$ & $\mathbf{0.167}$ & $\mathbf{0.163}$ \\
FDDM & $0.385$ & $0.527$ & $0.568$ \\
TDDM & $9.072$ & $22.110$ & $31.163$ \\
\tddmv{} & $2.388$ & $5.188$ & $7.201$ \\
IDM & $6.549$ & $5.426$ & $2.133$ \\
PSD & $0.334$ & $0.379$ & $0.357$ \\
\bottomrule
\end{tabular}
\end{table}

\subsection{Implementation Details}
\label{app:imp_detail}

For the trip, earthquake, and bent synthetic experiments, EFDM uses
discrete-time DDPM diffusion with the existence-weighted objective in
Eq.~\eqref{eq:efdm_loss}. QM9 uses continuous-time VP diffusion with
predictor--corrector sampling. The configurations are detailed below.

\paragraph{Checkpoint selection.}
For the trip, earthquake, and bent synthetic experiments, we select each
model's exponential moving average (EMA) checkpoint by the lowest
validation loss.

\paragraph{Diffusion schedules.}
For binary EFDM on trip, earthquake, and bent synthetic data, we use a
discrete DDPM with $T=1{,}000$ and the linear variance schedule
\[
 \beta_t=10^{-4}+\frac{t-1}{999}(0.02-10^{-4}),
 \qquad t=1,\ldots,1{,}000.
\]
Generation uses 1,000 reverse predictor steps without a corrector. The
binary EFDM sampler retains Gaussian variance $\beta_1$ in the final
reverse step. The other non-PSD methods also use 1,000 discrete
coordinate-diffusion steps; PSD uses its native 100-step
thinning/superposition sampler.
QM9 instead uses continuous-time VP diffusion with
$\beta(t)=0.1+19.9t$ for $t\in[0.001,1]$ during training. Its
predictor--corrector evaluation is described below.

\paragraph{TDDM deletion schedules.}
For the base TDDM on trip, earthquake, and bent synthetic data, we
follow the sample-independent rate design of~\citet{campbell2023trans}.
Let $n_{\max}^{\mathrm{train}}=\max_{S\in\mathcal D_{\mathrm{train}}}|S|$.
The shared forward deletion rate is calibrated to a cumulative intensity
of $n_{\max}^{\mathrm{train}}-1$ over the forward process. Its rate is
not adapted to each sample's initial count.

For \tddmv{} on these datasets, the forward deletion schedule has
$T=1{,}000$ steps. Given the initial cardinality $n$, its per-step
deletion rate is
\[
 \lambda_t(n)=
 \begin{cases}
  0, & 1\le t\le100\text{ or }901\le t\le1{,}000,\\
  \dfrac{\max(n-1,0)}{800}, & 101\le t\le900.
 \end{cases}
\]
Thus deletion is disabled for the first and last 10\% of steps. The
initial cardinality is observed during training and replaced by the
model's current prediction during generation.

For both variants, given an initial cardinality $n$, the forward count
is $n_t=\max(n-D_t,1)$, where
$D_t\sim\operatorname{Poisson}(\sum_{s=1}^{t}\lambda_s)$.
This truncation ensures that at least one point remains.

\paragraph{Point-set representation.}
For binary EFDM, a training point set with cardinality $n\le N$ is embedded in
$N$ slots.
We encode active and inactive slots with finite existence logits
corresponding to probabilities $1-\delta$ and $\delta$, respectively,
where $\delta\in(0,1/2)$.
We set $\delta=0.003$ for the trip, earthquake, and bent synthetic datasets.
This encoding preserves a strong separation between active and inactive
slots while keeping the logits finite:
\[
 u_i^{(0)}=
 \begin{cases}
  \log\dfrac{1-\delta}{\delta}, & \text{active},\\[4pt]
  -\log\dfrac{1-\delta}{\delta}, & \text{inactive}.
 \end{cases}
\]
The inactive coordinates are sampled from the observed points in the same
set.
This yields a fixed number of slots while encoding the original
cardinality through the existence variables.

\paragraph{Attribute encodings.}
For the shared-existence representation in Section~\ref{sec:typed},
finite log-ratio coordinates for categorical labels can be obtained
using a small amount of label smoothing. Active conditional labels
can be encoded as
\[
 \rho_{ik}^{(m)}=(1-\delta)\mathbf{1}\{a_i^{(m)}=k\}+\frac{\delta}{K_m}.
\]
Inactive slots use uniform conditional probabilities as auxiliary
placeholders.
The direct single-attribute parameterization used for QM9 and its
relation to these coordinates are described in
Appendix~\ref{app:attribute_parameterization}.

\paragraph{Maximum cardinality. }
In our experiments, we choose EFDM's maximum slot number $N$ based on the dataset cardinality range.
For the trip dataset, we use $N=400$, which is at least as large as the maximum observed cardinality.
We use $N=560$ for the earthquake dataset, $N=29$ for QM9 molecules, and $N=110$ for the bent synthetic benchmark.
Subject to available computational and memory budgets, $N$ can be set
as large as the task requires; it need not be close to the maximum
cardinality in the training data.

\paragraph{Model configurations.}
For the trip and earthquake datasets, EFDM, FDDM, TDDM, \tddmv{}, and PSD use Transformer-based architectures~\citep{vaswani2017attention} with the dimensions listed in Table~\ref{tab:app_model_configs_spatial}.
IDM uses a point-wise feedforward network, since it models points independently and does not use interactions among points.
TDDM and \tddmv{} use fused width 128 on both datasets.

\begin{table}[h]
	\centering
	\small
	\caption{
		Model configurations for the trip and earthquake experiments.
		EFDM, FDDM, TDDM, \tddmv{}, and PSD use Transformer-based architectures.
		IDM uses a point-wise MLP.
	}
	\label{tab:app_model_configs_spatial}
	\setlength{\tabcolsep}{2pt}
	\begin{tabular}{@{}llccccc@{}}
		\toprule
		Dataset & Method & Backbone & Emb. Dim & FF/Hidden Dim & Heads & Layers \\
		\midrule
		Trip & IDM & MLP & 128 & 256 & -- & 8 \\
		Trip & EFDM/FDDM/TDDM/\tddmv/PSD & Transformer & 128 & 256 & 4 & 8 \\
		Earthquake & IDM & MLP & 128 & 256 & -- & 8 \\
		Earthquake & EFDM/FDDM/TDDM/\tddmv/PSD & Transformer & 128 & 256 & 4 & 8 \\
		\bottomrule
	\end{tabular}
\end{table}
For QM9, EFDM uses an EGNN backbone~\citep{satorras2021en} and a
Transformer component~\citep{vaswani2017attention},
configured as shown in Table~\ref{tab:app_model_configs_qm9}.
Existence probabilities are used in the fusion branch but are not supplied as explicit inputs to the EGNN.
The model is trained with batch size 64 and learning rate $3\times 10^{-5}$.
All QM9 baseline results in the main paper are taken from the literature;
these methods are not retrained for the comparison.
% QM9 configuration: featgate=false; egnn_embedding_out_mode=1
% (the separate embedding output is not created or computed).
% TODO: add the parameter count and compute resources for the latest QM9 model.

\begin{table}[h]
	\centering
	\small
	\caption{
		Components of the EFDM model used for the QM9 molecule experiment.
	}
	\label{tab:app_model_configs_qm9}
	\begin{tabular}{lccc}
		\toprule
		Component & Hidden/Embedding Dim & Heads & Layers \\
		\midrule
		EGNN & 256 & -- & 9 \\
		Transformer & 256 & 8 & 4 \\
		\bottomrule
	\end{tabular}
\end{table}

\paragraph{QM9 training and label encoding.}
The QM9 EFDM training run spans 8,000 epochs, corresponding
to 800 million processed molecule instances. EMA used a half-life of
500,000 processed molecules with a ramp-up ratio of 0.05; at batch size
64, its asymptotic decay is $2^{-64/500000}\approx0.9999113$ per update.
In the direct categorical representation, the inactive reference logit
is fixed at $U_{i0}=0$. For an active slot, the observed atom-type logit
is $+1$ and every other type logit is $-1$; every type logit is $-1$
for an inactive slot. The binary-existence smoothing parameter
$\delta=0.003$ above does not define QM9 type labels.

\paragraph{QM9 checkpoint and evaluation.}
For the reported QM9 result, we use the EMA weights from the last training
epoch (epoch 8,000). We evaluate this checkpoint once by generating
10,000 molecules in batches of 64, rather than averaging over repeated
runs. Atom stability is $99.04\%$, molecule stability is $91.18\%$, and
validity is $96.38\%$.
Across six late EMA checkpoints (epochs 6,000--8,000), atom stability
ranges from $98.99\%$ to $99.10\%$, molecule stability from $90.23\%$
to $91.38\%$, and validity from $95.91\%$ to $96.38\%$; every checkpoint
exceeds the FDDM and TDDM scores on all three metrics.
We use predictor--corrector sampling.
With diffusion time normalized to $T=1$, the predictor step size is
$0.05$ for $t>0.5T$ and $0.001$ thereafter.
Corrector updates are enabled only for $0.003\le t<0.1T$, with five
Langevin updates per predictor step and a signal-to-noise ratio (SNR)
of $0.3$.
The sampler uses 510 predictor updates and 485 corrector updates,
for a total of 995 network function evaluations (NFE).
No noise is added in the final predictor update.

% The old trip runtime (5--10 hours) does not describe the new 60k-epoch
% trip experiment; its compute resources and runtime require verification.

\subsection{Single-Attribute Parameterization and Readout}
\label{app:attribute_parameterization}

For a single attribute with $K$ meaningful categories, the shared-existence
representation in Section~\ref{sec:typed} uses
$(x_i,u_i,v_{i2},\ldots,v_{iK})$, with $v_{i1}=0$.
An alternative parameterization fixes the logit of the inactive category
to zero and represents the entire categorical field directly:
\begin{equation}
 \begin{aligned}
 U_{i0}&=0,\qquad
 U_{ik}=\log\frac{\gamma_{ik}}{\gamma_{i0}},\quad k=1,\ldots,K,\\
 \boldsymbol{\gamma}_i
 &=\operatorname{softmax}(0,U_{i1},\ldots,U_{iK}).
 \end{aligned}
 \label{eq:efdm_attribute_logratio}
\end{equation}
Both parameterizations have $K$ free coordinates and represent every
categorical distribution in the simplex interior. The QM9 experiments
use the direct representation, jointly diffusing
$(x_i,U_{i1},\ldots,U_{iK})$.

\paragraph{Equivalence of the categorical fields.}
Suppress the slot index and write
$Z_v=\sum_{k=1}^{K}\exp(v_k)$, where $v_1=0$.
The map from shared-existence coordinates to direct logits is
\begin{equation}
 U_k=u+v_k-\log Z_v,\qquad k=1,\ldots,K.
 \label{eq:efdm_shared_logits}
\end{equation}
Indeed, $\sum_k\exp(U_k)=\exp(u)$, so
\begin{equation}
 \gamma_0=\frac{1}{1+\exp(u)}=1-\sigma(u),
 \qquad
 \gamma_k=\frac{\exp(u)}{1+\exp(u)}\frac{\exp(v_k)}{Z_v}
 =\sigma(u)\rho_k.
 \label{eq:efdm_parameterization_probabilities}
\end{equation}
Conversely, any $\boldsymbol{U}\in\mathbb{R}^{K}$ determines
\begin{equation}
 u=\log\sum_{k=1}^{K}\exp(U_k),\qquad
 v_k=U_k-U_1,\qquad
 \rho_k=\exp(U_k-u).
 \label{eq:efdm_single_attribute}
\end{equation}
These maps are inverses: substituting the latter expression gives
$\log Z_v=u-U_1$ and recovers every $U_k$ in
Eq.~\eqref{eq:efdm_shared_logits}.
Thus a categorical draw from $\boldsymbol{\gamma}$ has exactly the same
distribution as drawing presence from $\operatorname{Bernoulli}(\sigma(u))$
and, when present, drawing the attribute from
$\operatorname{Categorical}(\boldsymbol{\rho})$.

The two parameterizations represent the same categorical distributions
and probabilistic readout. For $K>1$, independent Gaussian diffusion in
these nonlinearly related coordinates generally yields different
dynamics and training objectives.

\paragraph{Deterministic readout in QM9.}
QM9 trains these direct logits with the fixed $\pm1$ encoding in
Appendix~\ref{app:imp_detail} and decodes the recovered full field by
argmax:
\begin{equation}
 \hat c_i=\operatorname*{arg\,max}_{k\in\{0,\ldots,K\}}\hat U_{ik},
 \qquad \hat U_{i0}=0.
 \label{eq:efdm_qm9_map}
\end{equation}
Category $0$ discards the slot; otherwise, the slot is retained with
attribute $\hat c_i$. With a consistent tie-breaking convention, the
corresponding MAP decision is identical in either parameterization
because the full categorical probabilities agree.
Away from ties, retention is equivalent to
\begin{equation}
 \max_k\hat U_{ik}>0
 \quad\Longleftrightarrow\quad
 \hat e_i\max_k\hat\rho_{ik}>1-\hat e_i.
 \label{eq:efdm_qm9_retention}
\end{equation}
This deterministic rule is generally different from either probabilistic
readout or thresholding the total existence probability at $1/2$.

\subsection{Evaluation Metric Details}
\label{app:metric_details}

Let $P$ and $Q$ denote the real and generated distributions over point sets
$S\subset\mathbb{R}^d$. Write $n=|S|$ for the cardinality of a point set
and $P_{\mathrm{card}},Q_{\mathrm{card}}$ for the corresponding marginal
distributions.
Let $\mathcal R=(R_i)_{i=1}^{r}$ and $\mathcal G=(G_j)_{j=1}^{g}$ be
the real test and generated collections, and let
$\mathcal T=(T_a)_{a=1}^{t}$ be the training collection.
Throughout, a collection contains point sets, rather than individual points.
The kernel estimators below require $r,g\ge2$ and positive bandwidths.

\paragraph{Cardinality and spatial summaries.}
Cardinality error is $W_1(\widehat P_{\mathrm{card}},\widehat Q_{\mathrm{card}})$, the one-dimensional
Wasserstein distance between the empirical count distributions.
For a nonempty set $S=\{x_i\}_{i=1}^{n}$, coordinate $a$, and moment
order $h$, define
\begin{equation}
 \bar x_a=\frac{1}{n}\sum_{i=1}^n x_{ia},
 \qquad
 m_{h,a}=\frac{1}{n}\sum_{i=1}^n(x_{ia}-\bar x_a)^h.
 \label{eq:spatial_moments}
\end{equation}
The coordinate-wise summaries are
\begin{equation}
 \mathrm{Mean}_a=\bar x_a,\quad
 \mathrm{Var}_a=m_{2,a},\quad
 \mathrm{Skew}_a=\frac{m_{3,a}}{m_{2,a}^{3/2}},\quad
 \mathrm{Kurt}_a=\frac{m_{4,a}}{m_{2,a}^{2}}-3.
 \label{eq:spatial_summaries}
\end{equation}
Variance divides by $n$, not $n-1$. Skew and Kurt are the uncorrected
standardized third central moment and Fisher excess kurtosis, respectively;
the formulas above define them when coordinate variance is positive.
For numerical stability, the reported implementation instead uses standardized
coordinates $z_i=(x_{ia}-\bar x_a)/(s_a+10^{-8})$ when the coordinate
standard deviation $s_a=\sqrt{m_{2,a}}$ is at least $10^{-8}$, and sets
all $z_i=0$ when $s_a<10^{-8}$. It then computes Skew as
$n^{-1}\sum_i z_i^3$ and Kurt as $n^{-1}\sum_i z_i^4-3$.
Thus coordinates with $s_a<10^{-8}$ contribute skewness $0$ and Fisher
excess kurtosis $-3$. Singleton sets remain in the spatial-summary
distributions.
The mean nearest-neighbor distance, for $n\ge2$, is
\begin{equation}
 \mathrm{NN1}(S)=\frac{1}{n}\sum_{i=1}^n
 \min_{j\ne i}\|x_i-x_j\|_2.
 \label{eq:nn1}
\end{equation}
Only observed points enter these summaries; inactive or padding slots
are excluded.
For a coordinate-wise summary $f$, the reported discrepancy is
\begin{equation}
 \mathcal W_f(\mathcal R,\mathcal G)
 =\frac{1}{d}\sum_{a=1}^{d}
 W_1\!\left(\frac{1}{r}\sum_{i=1}^{r}\delta_{f_a(R_i)},
             \frac{1}{g}\sum_{j=1}^{g}\delta_{f_a(G_j)}\right).
 \label{eq:feature_w1}
\end{equation}
This compares per-set feature distributions along each coordinate before
averaging the distances; it does not pool the individual points.
For NN1, the analogous formula has one scalar feature and no coordinate
average.

For the Earthquake and Trip experiments, empty generated proposals are
rejected and resampled until 4,096 nonempty sets are obtained per method
and split. All ten reported metrics, including Count $W_1$ and the
spatial-summary discrepancies in Eq.~\eqref{eq:feature_w1}, use this
same accepted generated collection. NN1 is set to zero for singleton
sets.
All 1,024 generated sets per method in the bent synthetic experiment
are also nonempty.

\paragraph{Sliced distance and normalization.}
For nonempty sets, spatial MMD uses the 50-projection sliced-$W_2$ distance
\citep{ludke2024unlocking}:
\begin{equation}
 D_{50}(S,S')=
 \left[\frac{1}{50}\sum_{\ell=1}^{50}
 W_2^2\bigl((\theta_\ell^\top)_\#\mu_S,
            (\theta_\ell^\top)_\#\mu_{S'}\bigr)\right]^{1/2},
 \quad
 \mu_S=\frac{1}{|S|}\sum_{x\in S}\delta_x.
 \label{eq:sw50}
\end{equation}
Here $\theta_\ell$ are random unit projection directions and $\mu_S$ is
the uniformly weighted empirical point measure.

Spatial-summary $W_1$ uses coordinates standardized with training-set
statistics for Earthquake and Trip.
For the bent synthetic benchmark, it uses the observed bent coordinates
without standardization or inverse transformation.
The MMD, calibrated MMD, and train-scale MMD evaluations reverse this
standardization, map the supplied spatial bounds to $[-1,1]^d$, and clip
coordinates to this domain.
JMMD and ICMMD use the bounded coordinates for Earthquake, but use
unclipped standardized coordinates for Trip.
The bent synthetic benchmark uses training-derived bounds and the same
bounded mapping and clipping as Earthquake for all MMD-family metrics.
Its structural diagnostics instead apply the inverse bending transform
to both real and generated points.
Their implementations do not normalize inputs internally.
Within each dataset and split, the same coordinate transformation is
applied to training, test, and generated sets for a given metric.
A three-seed aggregate is reported only when all three scores are finite.
The official PSD Wasserstein path draws projections without passing a
fixed seed to each distance call.
JMMD and ICMMD instead reuse the same integer projection seed for every
pair, ensuring the same 50 directions in a common ambient dimension.
For equal-cardinality nonempty sets, their distance has the equivalent form
\begin{equation}
 D_{50}(S,S')^2
 =\frac{1}{50n}\sum_{\ell=1}^{50}\sum_{i=1}^{n}
    (s_{\ell,(i)}-s'_{\ell,(i)})^2,
 \label{eq:equal_count_sw}
\end{equation}
where $s_{\ell,(i)}$ is the $i$th sorted projection of $S$ along
$\theta_\ell$.
Thus $\kappa_\sigma$ is a Gaussian kernel on the sorted projected-coordinate
representation within each cardinality.
The implementation sets $D_{50}(\emptyset,\emptyset)=0$ for these two
metrics; unequal-count pairs have zero joint kernel and do not require a
distance calculation.
Finitely many projections do not in general identify every distribution
of the original point sets.

\paragraph{MMD estimators.}
We use the MMD framework of~\citet{gretton2012kernel}, with a
Wasserstein-distance kernel for spatial point sets as in
PSD~\citep{ludke2024unlocking}. The metric labeled MMD uses the
unsquared sliced distance from Equation~\eqref{eq:sw50}:
\[
 k_\sigma(S,S')
 =\exp\!\left[-\frac{D_{50}(S,S')}{2\sigma^2}\right].
\]
For a kernel $k$, define the biased estimator, including self-pairs, by
\begin{equation}
 \widehat D_{v,k}(\mathcal R,\mathcal G)
 =\frac{1}{r^2}\sum_{i,i'=1}^{r}k(R_i,R_{i'})
 +\frac{1}{g^2}\sum_{j,j'=1}^{g}k(G_j,G_{j'})
 -\frac{2}{rg}\sum_{i=1}^{r}\sum_{j=1}^{g}k(R_i,G_j).
 \label{eq:mmd_v}
\end{equation}
MMD reports $\sqrt{\widehat D_{v,k_\sigma}}$.
For its default bandwidth, form the concatenation (with multiplicities)
\begin{equation}
 \begin{aligned}
 \mathcal D_{\mathrm{eval}}
 &=\bigl(D_{50}(R_i,R_{i'})\bigr)_{i,i'}
 \mathbin{\Vert}\bigl(D_{50}(G_j,G_{j'})\bigr)_{j,j'}
 \mathbin{\Vert}\bigl(D_{50}(R_i,G_j)\bigr)_{i,j},\\
 \sigma_{\mathrm{eval}}&=\operatorname{median}\mathcal D_{\mathrm{eval}}.
 \end{aligned}
 \label{eq:mmd_eval_sigma}
\end{equation}
Each within-collection block contains all ordered pairs, including its
diagonal, and the cross block is included once.
This is the bandwidth rule of the official PSD implementation
\citep{ludke2024unlocking}, which may yield a different bandwidth for each
generated collection.
Calibrated MMD retains the distance routine, kernel $k_\sigma$, and square-root
biased estimator, but fixes its bandwidth from the training collection:
\begin{equation}
 \sigma_{\mathrm{train}}=
 \operatorname{median}\{D_{50}(T_i,T_j):1\le i,j\le t\}.
 \label{eq:mmd_train_sigma}
\end{equation}
The median includes all $t^2$ ordered pairs, including self-pairs, and is
fixed across methods for a given training split without using test or
generated samples to select it.

For train-scale MMD, let $m=\sigma_{\mathrm{train}}$ denote the same
median training distance and instead use
\begin{equation}
 k_{\mathrm{train\mbox{-}scale}}(S,S')
 =\exp\!\left[-\frac{D_{50}(S,S')}{2m}\right].
 \label{eq:mmd_train_scale}
\end{equation}
The metric still reports the square root of the biased MMD$^2$ estimator,
but its kernel denominator is $2m$, rather than $2m^2$ as in calibrated
MMD. Equivalently, the squared kernel bandwidth is $m$.
These are distinct metrics and are labeled separately.
The implementation uses 50 fresh random projections per ordered pair
and full real--real, generated--generated, and cross-collection grids.
Both MMD variants use all 4,096 generated sets against 300 test sets
for Earthquake and 91 test sets for Trip. The bent synthetic evaluation uses all 1,024
generated and 2,000 test sets.

The unbiased squared-MMD estimator removes only within-collection
self-pairs~\citep{gretton2012kernel}:
\begin{equation}
 \widehat D_{u,k}(\mathcal R,\mathcal G)
 =\frac{\sum_{i\ne i'}k(R_i,R_{i'})}{r(r-1)}
 +\frac{\sum_{j\ne j'}k(G_j,G_{j'})}{g(g-1)}
 -\frac{2}{rg}\sum_{i,j}k(R_i,G_j).
 \label{eq:mmd_u}
\end{equation}
For joint cardinality and spatial comparisons, define the product kernel
\citep{long2017joint}
\begin{equation}
 K_\sigma((n,S),(n',S'))=
 \mathbf{1}_{\{n=n'\}}\kappa_\sigma(S,S'),
 \qquad
 \kappa_\sigma(S,S')=
 \exp\!\left[-\frac{D_{50}(S,S')^2}{2\sigma^2}\right].
 \label{eq:jmmd_kernel}
\end{equation}
JMMD reports $\widehat D_{u,K_\sigma}$, a signed estimate of
$\mathrm{MMD}_{K_\sigma}^2(P_{\mathrm{card},S},Q_{\mathrm{card},S})$.
The denominators are the total pair counts, not the numbers of pairs with
equal cardinality.
For a fixed kernel and independent samples, this is an unbiased estimate
of the population squared discrepancy.
It is reported with its sign, without a square root or clipping.
Negative estimates can occur through finite-sample fluctuations; their
sign alone does not establish statistical indistinguishability.

\paragraph{Training bandwidth for JMMD and ICMMD.}
Let
\begin{equation}
 \begin{aligned}
 \mathcal P_{\mathrm{eq}}
 &\subseteq\{(a,b):1\le a<b\le t,\ |T_a|=|T_b|\},\\
 \sigma_{\mathrm{eq}}
 &=\operatorname{median}\{D_{50}(T_a,T_b):(a,b)\in\mathcal P_{\mathrm{eq}}\}.
 \end{aligned}
 \label{eq:equal_count_sigma}
\end{equation}
All distinct unordered equal-count training pairs are used if there are
at most 20,000; otherwise 20,000 are sampled uniformly without replacement.
The pairs are pooled across counts, so count groups are weighted by their
numbers of pairs in this bandwidth calculation.
Self-pairs are excluded.
The resulting bandwidth is shared by JMMD and ICMMD and fixed across
methods for the same split and projection seed.
Sensitivity evaluations at $0.5\sigma_{\mathrm{eq}}$ and
$2\sigma_{\mathrm{eq}}$ can reuse the same distances.
This convention differs from the all-ordered-pair training median used
for calibrated MMD.
The reported Earthquake and Trip evaluations use the training equal-count
bandwidth with a fixed projection seed, shared across methods within a split.
For Earthquake, the bandwidths are approximately 0.1730, 0.1750, and
0.1692 across the three split/training seeds.

\paragraph{Joint versus conditional discrepancies.}
The product-kernel construction follows joint mean embeddings
\citep{long2017joint}.
Write $P_n=P(S\mid |S|=n)$ and $Q_n=Q(S\mid |S|=n)$ wherever the
corresponding count probabilities are positive.
If $p_n=P_{\mathrm{card}}(n)$, $q_n=Q_{\mathrm{card}}(n)$, and
$\mu_{P_n},\mu_{Q_n}$ are the spatial
kernel mean embeddings within cardinality $n$, then
\begin{equation}
 \mathrm{JMMD}^2(P,Q)
 =\sum_n\|p_n\mu_{P_n}-q_n\mu_{Q_n}\|_{\mathcal H_n}^{2}.
 \label{eq:jmmd_decomposition}
\end{equation}
A weighted embedding is defined to be zero when its count probability is
zero, so this expression does not require a conditional law at an absent
count.
It therefore compares both the count probabilities and the associated
spatial distributions.
ICMMD instead integrates the conditional discrepancies against the real
cardinality marginal:
\begin{equation}
 \mathrm{ICMMD}^2(P,Q;P_{\mathrm{card}})
 =\sum_{n:P_{\mathrm{card}}(n)>0}P_{\mathrm{card}}(n)\,
   \mathrm{MMD}_{\kappa_\sigma}^2(P_n,Q_n),
 \label{eq:icmmd_target}
\end{equation}
This population target requires $Q_n$ to be defined on the support of
$P_{\mathrm{card}}$.
This is a discrete-cardinality adaptation of integrated conditional MMD
\citep{yan2024conditional}; that reference studies continuous conditioning
variables and kernel-smoothed estimators.
Our exact-count eligibility, weighting, and penalty rules are specified
below.

\paragraph{Reported ICMMD score and coverage.}
Write $\mathcal R_n=\{R_i:|R_i|=n\}$,
$\mathcal G_n=\{G_j:|G_j|=n\}$, $r_n=|\mathcal R_n|$, and
$g_n=|\mathcal G_n|$.
Using the implementation's default minimum real group size of two, define
\begin{equation}
 \mathcal E=\{n:r_n\ge2\},\qquad
 \mathcal C=\{n\in\mathcal E:g_n\ge2\},\qquad
 w_n=\frac{r_n}{\sum_{a\in\mathcal E}r_a}.
 \label{eq:icmmd_eligibility}
\end{equation}
Eligibility and weights depend only on the real test counts.
For a collection $A=(A_i)_{i=1}^{s}$ with $s\ge2$, let
\begin{equation}
 U_\sigma(A)=\frac{2}{s(s-1)}\sum_{i<i'}\kappa_\sigma(A_i,A_{i'}).
\end{equation}
The reported score is
\begin{equation}
 \widehat{\mathrm{ICMMD}}^{\,2}
 =\sum_{n\in\mathcal E}w_n\widehat\Delta_n,
 \qquad
 \widehat\Delta_n=
 \begin{cases}
 \widehat D_{u,\kappa_\sigma}(\mathcal R_n,\mathcal G_n),&n\in\mathcal C,\\
 1+U_\sigma(\mathcal R_n),&n\in\mathcal E\setminus\mathcal C.
 \end{cases}
 \label{eq:icmmd_estimator}
\end{equation}
The second case applies to both absent generated counts and generated
groups containing only one set.
Since $0\le\kappa_\sigma\le1$, $1+U_\sigma(\mathcal R_n)$ is a
conservative upper bound for a covered group's squared-MMD estimate.
Using it as a penalty prevents insufficient generated support from being
silently omitted.
The complete restricted and penalized score is not an unbiased estimator
of the unrestricted population target.
Covered-group estimates retain their signs without clipping.

To make the penalty's contribution visible, the implementation also returns
\begin{align}
 \mathrm{EligibleMass}&=\frac{\sum_{n\in\mathcal E}r_n}{r},
 &\mathrm{Coverage}&=\sum_{n\in\mathcal C}w_n,\\
 \widehat{\mathrm{ICMMD}}_{\mathrm{covered}}^{\,2}
 &=\frac{\sum_{n\in\mathcal C}w_n\widehat\Delta_n}
          {\sum_{n\in\mathcal C}w_n}.
 \label{eq:icmmd_coverage}
\end{align}
The main score and coverage are normalized by eligible real mass;
the covered score is normalized by covered real mass.
The covered score is undefined when coverage is zero, and the evaluation
is undefined if no real count is eligible.
% The supplied code returns 0 for an empty eligible collection; this must
% be treated as unavailable, not as evidence of a perfect distribution match.
% A missing equal-count training pair or zero median bandwidth also requires
% an explicit unavailable result or a separately documented fallback rule.

% Keep the final appendix discussion together before the full-page figures.
\begingroup
\setlength{\parskip}{0.25pc}
\subsection{Spatial evaluation protocols and supplementary results}
\label{app:spatial_evaluation}

\paragraph{Runs and checkpoint selection.}
The trip and earthquake comparisons use three split/training seeds.
Within a dataset and seed, all methods share the
same data split; changing the seed changes both the split and training
randomness. Runs are trained for 60,000 epochs, and each reported
checkpoint is selected by the lowest validation loss evaluated with EMA
weights. Test metrics are not used to select checkpoints.
Earthquake models use Adam~\citep{kingma2015adam}.
EFDM and FDDM use batches of 64 point sets, TDDM and \tddmv{} use
32 sets, PSD uses 128 sets, and IDM uses 2,048 individual points.
The learning rate is $10^{-4}$ with zero weight decay except for PSD,
which uses $10^{-3}$ and weight decay $10^{-4}$, with gradient clipping
at 2. All Earthquake methods use EMA decay 0.9999 per update; the
lowest-loss EMA checkpoint is selected from validation evaluations
every 300 epochs (200 candidates).

For trip, each split contains 268 training, 89 validation, and 91 test
point sets. Each day's point set is subsampled to approximately 10\%
of its points.
The density filter and coordinate normalization are fitted on the
training split.
Trip uses an EMA half-life of 100 epochs for
EFDM, IDM, and PSD, and 500 epochs for FDDM, TDDM, and
\tddmv{}, with ramp-up parameter 0.2.
Trip uses Adam~\citep{kingma2015adam} with learning rate $10^{-4}$ and
zero weight decay for non-PSD methods; PSD uses learning rate
$10^{-3}$ and weight decay $10^{-4}$. Training batches are 64 sets
for EFDM and FDDM, 32 sets for TDDM and \tddmv{},
128 sets for PSD, and 2,048 points for IDM.

\paragraph{Sampling and aggregation.}
Trip uses 4,096 generated point sets per method and seed against all
91 test sets, sampled in batches of 32 from the same selected EMA
checkpoints. Its samplers use 1,000 predictor steps for non-PSD
methods and PSD's native 100 steps, without a corrector. Earthquake
uses 4,096 generated sets against all 300 test sets for all ten
main-table metrics, including both MMD variants.
Both datasets report mean $\pm$ sample standard deviation across three
split/training seeds (divisor $3-1$), computed directly from unrounded
per-seed metric values. These standard deviations reflect split and
training variability; they are not standard errors or confidence
intervals. Training budgets are matched by epoch count, not by
computational cost.

The implementation names \texttt{mask}, \texttt{nogroup},
\texttt{jump}, and \texttt{jump\_adaptive} correspond to FDDM, IDM,
TDDM, and \tddmv{}, respectively. FDDM and IDM sample cardinalities
from a distribution fitted to training counts; the other methods
generate their own cardinalities.

\paragraph{Trip MMD saturation.}
All ten Trip metrics use the same 4,096 generated sets against all
91 test sets. MMD scores lie close to the diagonal-kernel value
$\sqrt{1/4096+1/91}\approx0.105987$, consistent with saturation.
The small differences in this column are not interpreted as evidence
of better generation quality. Both MMD and train-scale MMD remain in
the main table; they use the distinct bandwidth conventions in
Appendix~\ref{app:metric_details}.

\subsection{Additional Visualizations}
This section provides additional visualizations that complement the quantitative results in the main paper. 
The figures show molecule examples for QM9, together with cardinality summaries and point-set examples for the Earthquake and Trip datasets.

\paragraph{QM9 molecules.}
Figure~\ref{fig:qm9_examples} compares molecules from QM9 with
EFDM-generated samples with matching atom counts.
The examples include both chain-like and cyclic configurations.

\begin{figure}[t]
	\centering

	\includegraphics[width=\textwidth]{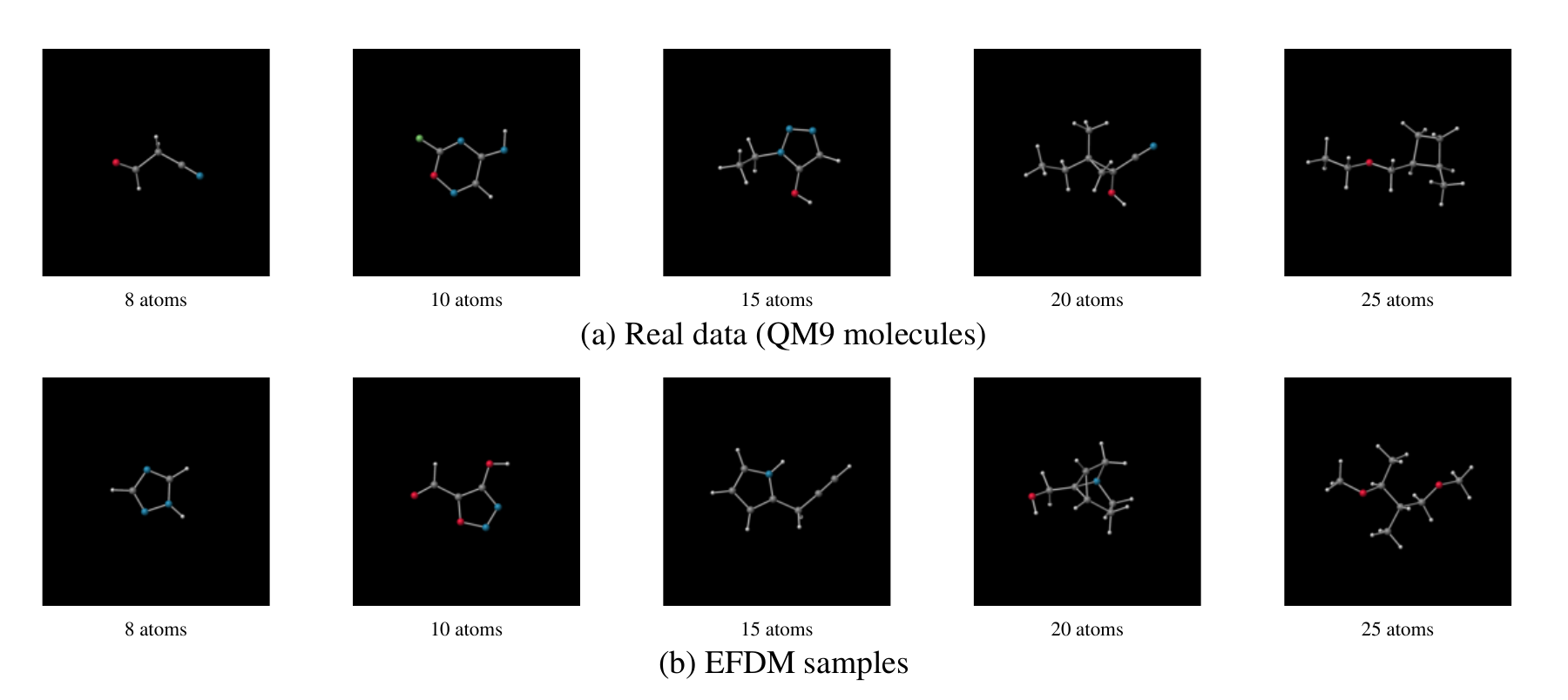}

	\caption{
		Examples of molecules from the QM9 dataset (top) and samples generated by EFDM (bottom).
		From left to right, each column shows molecules with 8, 10, 15, 20, and 25 atoms, respectively.
		Bonds are inferred from interatomic distances and are shown only for visualization.
	}
	\label{fig:qm9_examples}
\end{figure}

\paragraph{Spatial datasets.}
Figure~\ref{fig:spatial_count_coverage} summarizes nonempty generated
collections for Earthquake and Trip, generated without a corrector
and summarized over three split/training seeds.
TDDM's count distribution is shifted toward larger sets on both
datasets, while \tddmv{} concentrates on a narrower range than the real
data. The coverage panels quantify whether individual eligible counts
are represented, complementing the overall count-distribution comparison.

Figures~\ref{fig:eq_samples} and~\ref{fig:trip_plot} show point-set
examples in training-defined count ranges. Annotated cardinalities
show both matched counts and remaining differences within a range;
the high-count examples do not all have equal cardinality.
Blank panels indicate that the displayed collection contains no sample
in the corresponding range.

\begin{figure}[p]
    \centering
    \includegraphics[width=\linewidth]{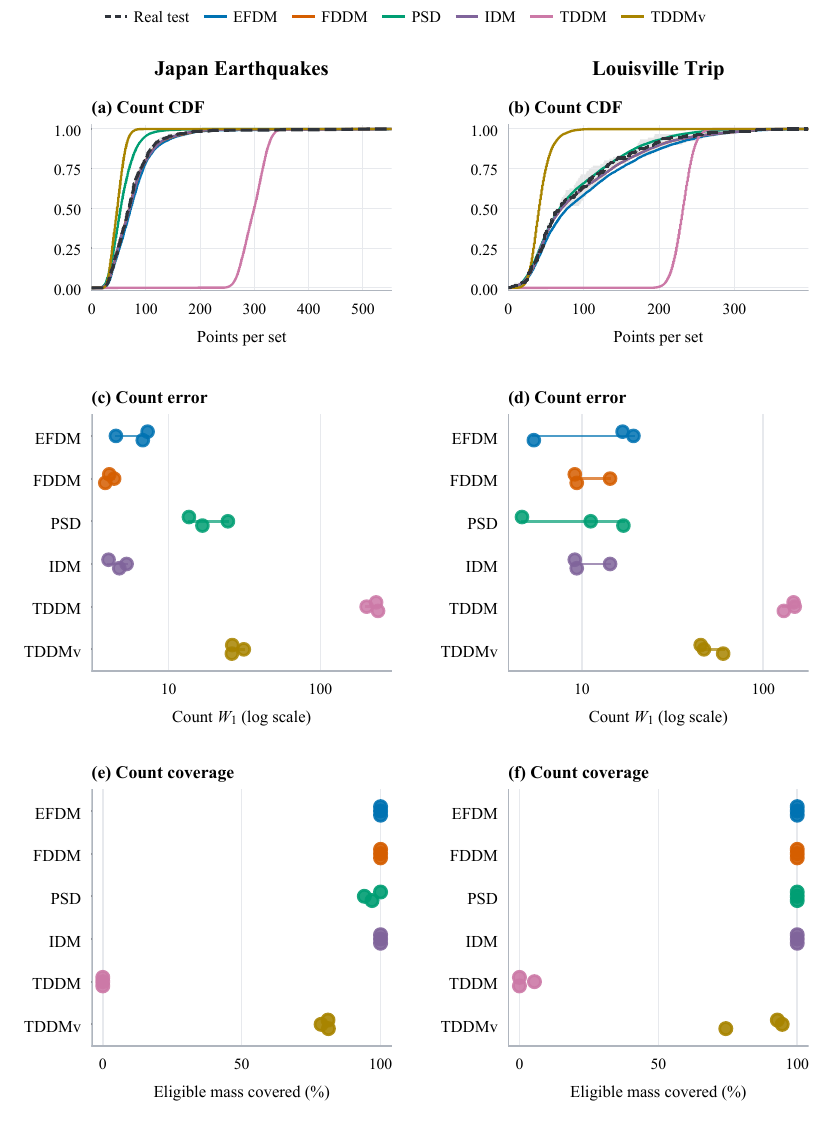}
    \caption{Cardinality summaries for Earthquake (left) and Trip (right),
    using 4,096 nonempty generated sets per method and split.
    Rows show count CDFs, count $W_1$, and coverage.
    CDFs are averaged over three splits; gray bands show the real split
    range. Dots denote individual splits.
    Coverage is the fraction of eligible test mass at counts with at
    least two generated sets; eligibility requires at least two real
    test sets at that count. Eligible mass is 86--87\% for Earthquake
    and 31--41\% for Trip.}
    \label{fig:spatial_count_coverage}
\end{figure}

\begin{figure}[p]
    \centering
    \includegraphics[width=\linewidth]{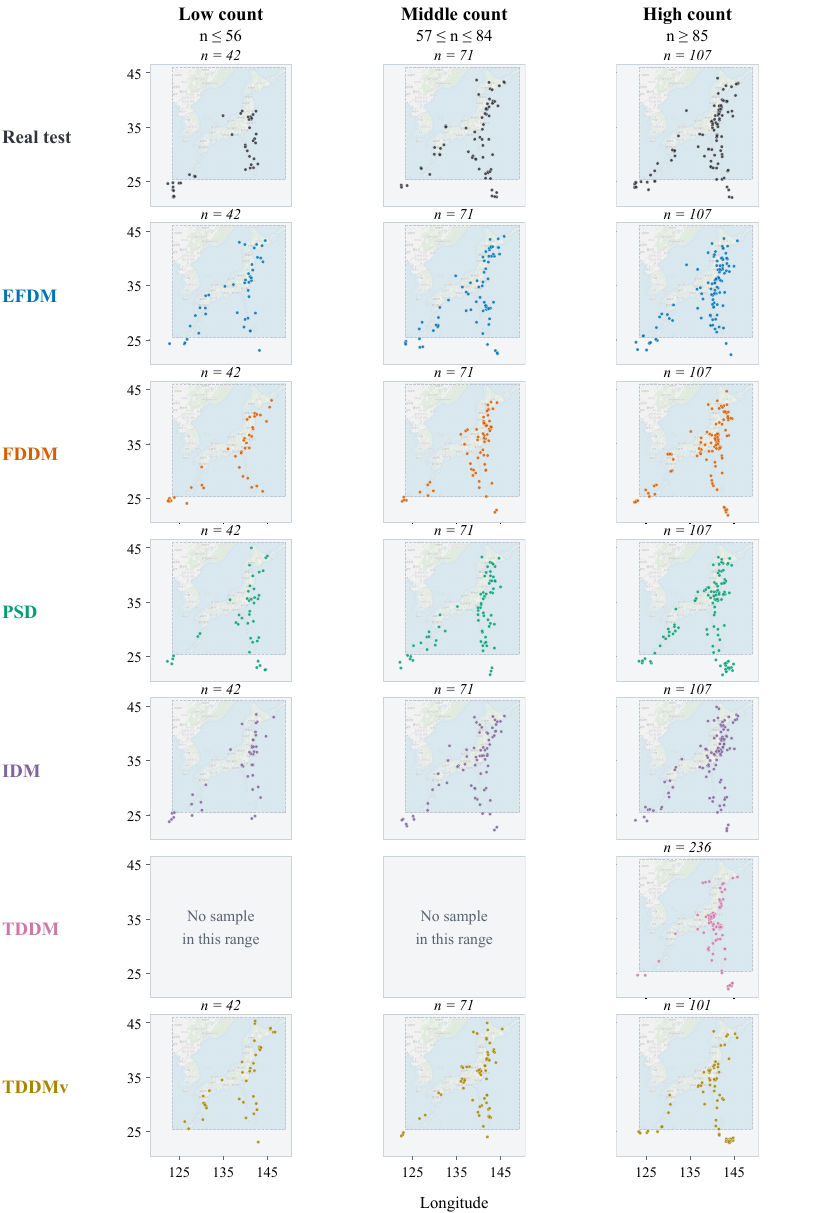}
    \caption{Japan Earthquakes: point-set examples with methods in rows
    and three training-defined count ranges in columns.
    Annotated counts show differences within each range.
    All panels share longitude--latitude axes; dashed boxes mark the
    map-background extent. Blank panels denote no sample in that range
    within the displayed collection.}
    \label{fig:eq_samples}
\end{figure}

\begin{figure}[p]
    \centering
    \includegraphics[width=\linewidth]{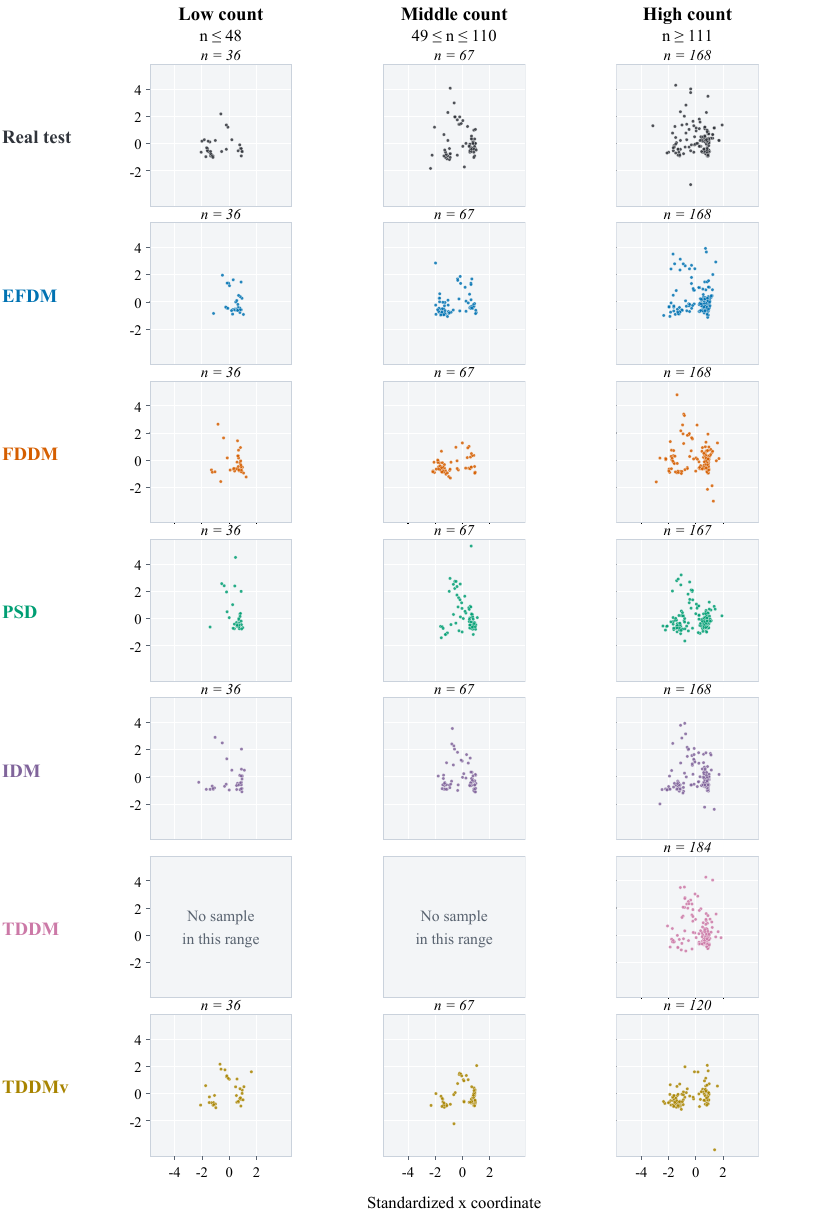}
    \caption{Louisville Trip: pickup-location sets with methods in rows
    and three training-defined count ranges in columns.
    Annotated counts show differences within each range.
    Coordinates use training standardization, shared axes, and no
    per-set alignment. Blank panels denote no sample in that range
    within the displayed collection.}
    \label{fig:trip_plot}
\end{figure}

\par
\endgroup

\end{document}